\documentclass[journal]{IEEEtran}
\ifCLASSINFOpdf
\else
\fi
\usepackage{threeparttable} 
\usepackage{amsmath}
\usepackage{bm}
\usepackage{multirow}
\usepackage{url}
\usepackage{lipsum}
\usepackage{graphicx}
\usepackage{color}
\ifCLASSOPTIONcompsoc
 \usepackage[caption=false,font=normalsize,labelfont=sf,textfont=sf]{subfig}
\else
 \usepackage[caption=false,font=footnotesize]{subfig}
\fi
\usepackage{}

\begin{document}
%
\title{Theme-Aware Aesthetic Distribution Prediction With Full-Resolution Photographs}
%
%
%

\author{Gengyun~Jia,
        Peipei~Li,
        and~Ran~He$^*$,~\IEEEmembership{Senior~Member,~IEEE}
\thanks{This work is partially funded by, National Natural Science Foundation of China (Grant No. U21B2045), National Natural Science Foundation of China (Grant No. U20A20223), and Beijing Nova Program (Grant No. Z211100002121106) (\textit{$*$Corresponding author: Ran He})}
\thanks{G. Jia is with the School of Artificial Intelligence, University of Chinese Academy of Sciences, Beijing 100049, China, and with the National Laboratory of Pattern Recognition, Center for Research on Intelligent Perception and Computing, Institute of Automation, Chinese Academy of Sciences, Beijing 100190, China (e-mail: gengyun.jia@cripac.ia.ac.cn).}
\thanks{P. Li is with the  School of Artificial Intelligence, Beijing University of Posts and Telecommunications, Beijing 100876, China (e-mail: lipeipei@bupt.edu.cn).}
\thanks{R. He is with the National Laboratory of Pattern
Recognition, Center for Research on Intelligent Perception and Computing,
CAS Center for Excellence in Brain Science and Intelligence Technology,
Institute of Automation, Chinese Academy of Sciences, Beijing 100190,
China, and with the University of Chinese Academy of Sciences,
Beijing 100049, China (e-mail: rhe@nlpr.ia.ac.cn).}
\thanks{The first two authors contribute equally.}}

%
%

\markboth{IEEE TRANSACTIONS ON NEURAL NETWORKS AND LEARNING SYSTEMS,~Vol.XX, No.xx, JANUARY, 2021}%
{Shell \MakeLowercase{\textit{et al.}}: Bare Demo of IEEEtran.cls for IEEE Journals}
%



\maketitle

\begin{abstract}

Aesthetic quality assessment (AQA) is a challenging task due to complex aesthetic factors. Currently, it is common to conduct AQA using deep neural networks that require fixed-size inputs. Existing methods mainly transform images by resizing, cropping, and padding or employ adaptive pooling to alternately capture the aesthetic features from fixed-size inputs. However, these transformations potentially damage aesthetic features. To address this issue, we propose a simple but effective method to accomplish full-resolution image AQA by combining image padding with region of image (RoM) pooling. Padding turns inputs into the same size. RoM pooling pools image features and discards extra padded features to eliminate the side effects of padding. In addition, the image aspect ratios are encoded and fused with visual features to remedy the shape information loss of RoM pooling. Furthermore, we observe that the same image may receive different aesthetic evaluations under different themes, which we call theme criterion bias. Hence, a theme-aware model that uses theme information to guide model predictions is proposed. Finally, we design an attention-based feature fusion module to effectively utilize both the shape and theme information. Extensive experiments prove the effectiveness of the proposed method over state-of-the-art methods.

\end{abstract}


%

\section{Introduction}
%
%
%
%

\begin{figure*} 
    \centering
  \subfloat[Original\label{1a}]{%
       \includegraphics[height=1.2in]{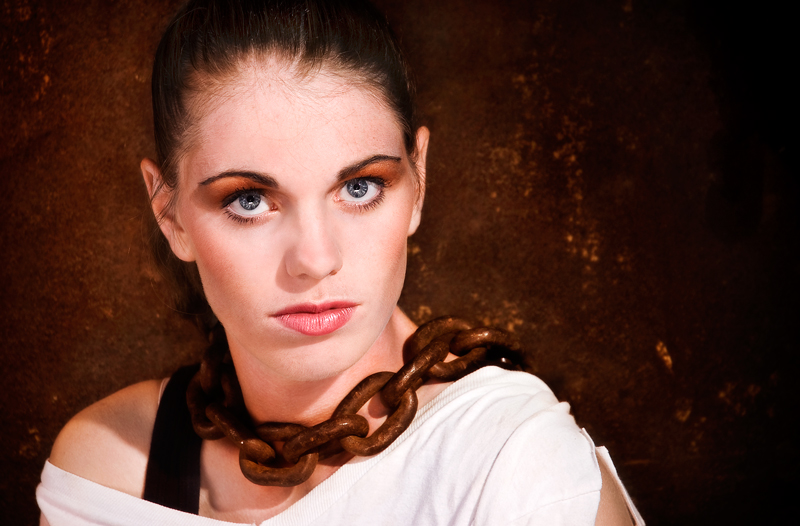}}
    \hfill
  \subfloat[Crop\label{1b}]{%
        \includegraphics[height=1.2in]{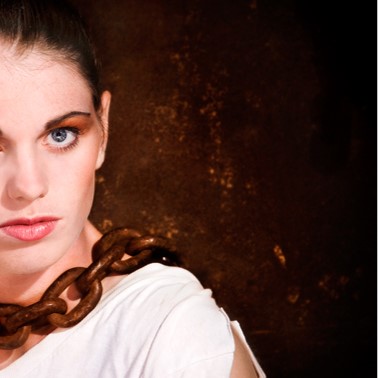}}
    \hfill
  \subfloat[Pad\label{1c}]{%
        \includegraphics[height=1.2in]{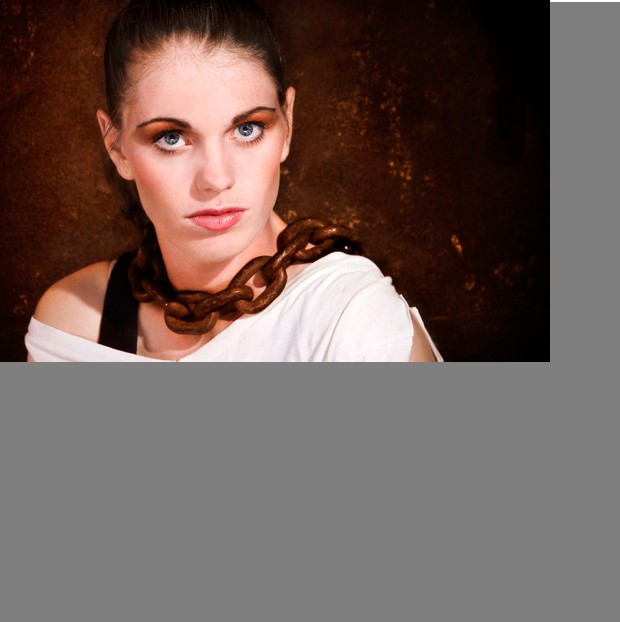}}
    \hfill
  \subfloat[Resize\label{1d}]{%
        \includegraphics[height=1.2in]{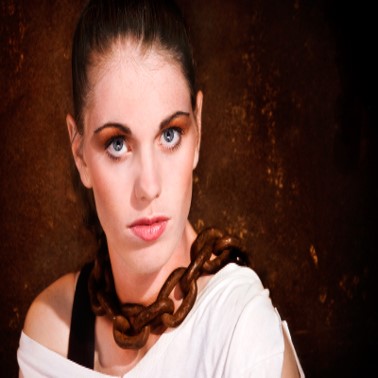}}
  \caption{Examples of transformations to fix the image size. (a): The original image is evaluated with a high average score (5.94). (b): Cropping destroys both the image layout and the object integrity. (c): The additional padded regions will confuse the network since the padded areas are different on different images. (d): The resizing operation warps the image and introduces noise.}
  \label{traditionaltransform} 
\end{figure*}

\IEEEPARstart{P}{hoto} aesthetic quality assessment (AQA) is an interesting task with wide applications, such as retrieving photos of high aesthetic quality and guiding aesthetic-driven image enhancement \cite{deng2018aesthetic}. However, the subjectivity and diversity of human assessment make the task complex. Many efforts \cite{ke2006design, tang2013content} have been made to leverage general human criteria, such as ``the rule of thirds", color harmony, and depth of field. Some researchers also directly introduced generic features widely adopted in pattern recognition \cite{ke2006design,tang2013content,bhattacharya2010framework,dhar2011high,su2011scenic,li2010towards}. Although the handcrafted features are carefully designed, they still suffer from limited representation ability.

With the development of deep learning methods \cite{he2016deep, szegedy2016rethinking} and the collection of large-scale databases for aesthetic analysis \cite{murray2012ava}, methods based on deep neural networks (DNNs) are widely employed in AQA. However, the size of inputs needs to be fixed for traditional DNNs. To handle the various sizes of real-world images, three types of image transformations are widely adopted. The first method that directly resizes different images is the most common solution, but considerable side effects, including distortion, blur, and artifacts are harmful to AQA, as shown in Fig.~\ref{1d}. The second method crops fixed-size patches from the original images \cite{Lu2015DeepMA, Ma2017ALampAL}. Although the original local resolution can be maintained, its damage to the image global structure and the layout information is irreversible. As Fig.~\ref{1b} shows, even the completeness of the image object is destroyed. The third method that pads images to the same size confuses the learning model because the shape and size of padded regions are different in different images, as shown in Fig.~\ref{1c}. In addition to image transformation methods, some works employ special networks such as fully convolutional networks (FCNs) \cite{long2015fully} and spatial pyramid pooling (SPP) \cite{he2015spatial}. However, these architectures only relax the constraint from one uniform input size to predefined multiple input sizes to adapt to the deep learning tools \cite{fang2018image,apostolidis2019image,cui2018distribution}. Obviously, it is insufficient to reflect the diversity of real-world image sizes with only a few predefined numbers. There are two solutions to absolutely avoid image transformations with these architectures. One solution is two-stage training without fine-tuning the feature extractor. The performance of this method is significantly limited by the insufficient feature extraction ability, and the training process is complicated. The other solution is to set the training batch size to one, but this results in unstable and inefficient training processes and worsens the performance \cite{apostolidis2019image}.

To address the conflict between the uniform input size constraint and diverse sizes of real-world images, we propose to combine padding with region of interest (RoI) pooling \cite{girshick2015fast}. RoI pooling is used to downsample features in a local region called an RoI in feature maps. In our model, we first pad the original images to a predefined size. The inputs, therefore, become the combination of an image region and a padded region, and the image region is regarded as the RoI. Then, the RoI pooling layer in the networks will only pool features inside the image region, leaving the features from the padded region unused. We call the RoI used in our model the RoM (Region of iMage) to distinguish it from the traditional RoI. For simplicity, this method is called padding with RoM pooling (PRP). Through this procedure, we finally achieve arbitrary input sizes with arbitrary batch sizes (as long as the GPU memory can hold) but introduce no sampling noise or useless information from padding pixels. To the best of our knowledge, this approach is the first method in AQA that supports end-to-end batch training on full-resolution images. Despite the advantages, the PRP module still neglects some useful information. For example, the original shape information is lost since the shapes of the RoM pooling outputs are the same. This information loss may cause performance degradation because image shapes influence human aesthetic perception \cite{zeng2020grid, tu2020image}. To remedy this shortcoming, we propose a shape-aware (SA) module to utilize the lost shape information. Specifically, the original image aspect ratios are firstly discretized and turned into one-hot codes. Then, we extract the shape features from the one-hot shape codes through fully connected layers. Finally, the shape features and the image features are fused to predict the aesthetic qualities.

\begin{figure} 
    \centering
    \includegraphics[width=0.95\linewidth]{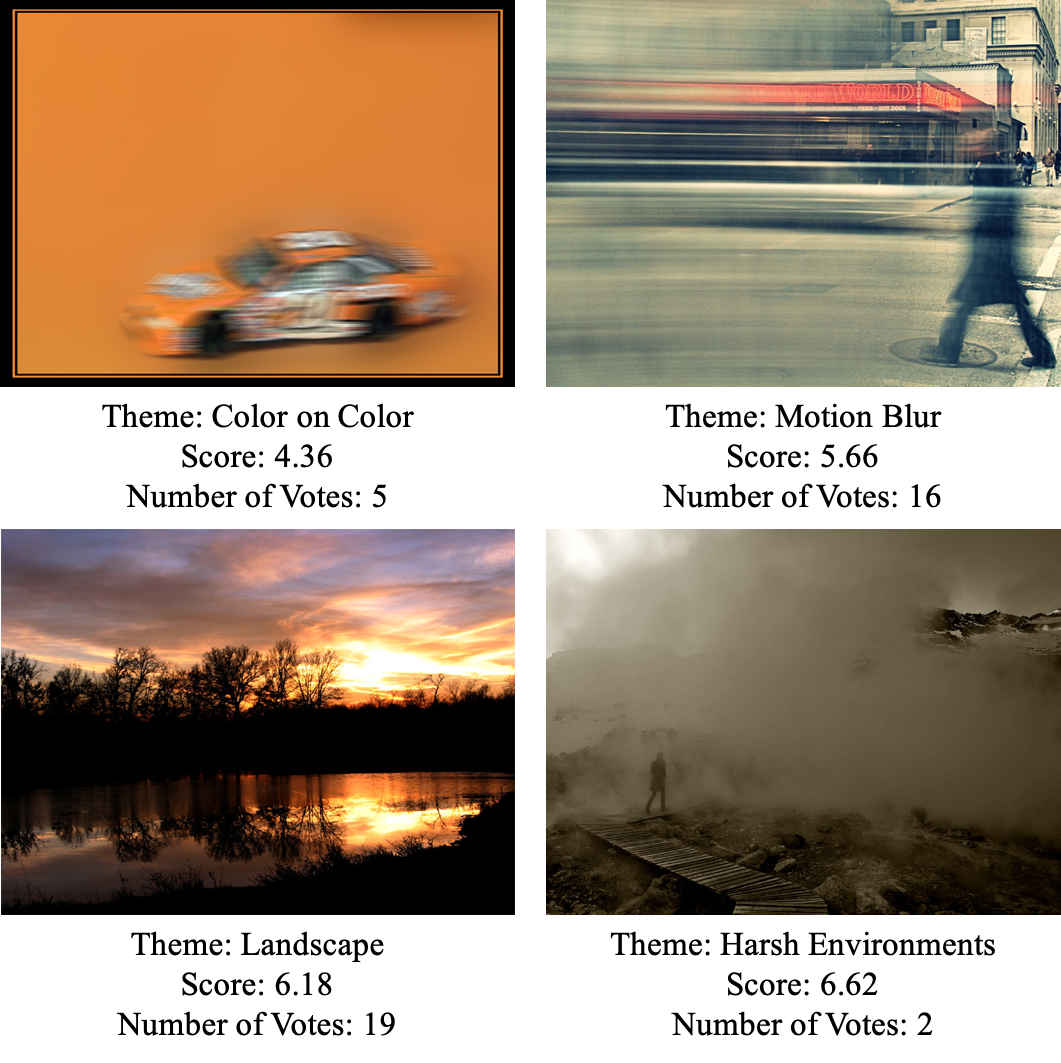}
    \caption{Examples of the theme criterion bias. We asked 21 people to compare the images in each row and voted for better images without knowing the corresponding themes. The "Number of Votes" gives the voting results. The images in the first row are both heavily blurred. We can see that the second image with a higher score also obtains more votes. However, in the second row, although the right image looks less attractive and receives significantly fewer votes, it obtains a higher score due to its fits to the theme.}
  \label{themecriterion} 
\end{figure}

The AQA criteria may be influenced by many factors \cite{tang2013content,ren2017personalized}. We find that a factor may significantly change the AQA criteria in the widely used AVA dataset. We call this factor a ``theme''. Specifically, the AVA dataset is collected from the website Dpchallenge.com which regularly holds photographic challenges with different themes. Photos taken by challenge participants are submitted to one of the challenges and evaluated by website users. There are some rules for the users to evaluate the challenge entries. One of the important rules is ``consider the challenge topics when voting, and adjust your score accordingly'' \cite{VotingGuidelines}. In other words, the photos are evaluated under the criteria influenced by the challenge themes. We use some typical examples to show how themes influence aesthetic criteria. We select two pairs of images from the AVA dataset and ask some people to vote on them. Specifically, 21 people are asked to select the most attractive image in each pair without knowing their corresponding themes (we try to keep the other conditions, such as the presentation method, the same as for website users). In Fig.~\ref{themecriterion}, the corresponding themes, ground-truth average scores, and voting results are given below the images. Each pair of images are shown in one row. The two images in the first row are both highly blurred. The right image obtains a significantly higher score than the left image. One possible reason is that the right image comes from the theme ``motion blur'', in which blurring is regarded as a good feature. The images in the second row are all of natural scenes. Although the left image that belongs to the theme "Landscape" is considered more beautiful by most users who do not know the themes, the score for it is lower than that for the right image, possibly because the content and the style of the right image better match the theme ``harsh environments''. In summary, themes are the challenge topics that photos need to match. If images are the entries of the challenges from dpchallenge.com, predicting their aesthetic qualities without the corresponding themes may bring about inaccurate results because of criterion bias. To address this issue, it is natural to introduce the challenge themes in AQA. The theme information is encoded and combined with the extracted visual features. With the help of theme features, visual features can be extracted and evaluated adaptively, which makes the assessment consistent with the human evaluation process. 

As described before, two extra features need to be fused with image visual features in our model. One extra fearture is the aspect ratio feature aiming to remedy the shape information loss of the PRP module, and the other extra feature is the theme feature aiming to introduce theme criterion bias. To better fuse different features, we propose an attention-based feature fusion module. In this module, the relations between the extra features and the visual features are captured with the help of the attention mechanism \cite{vaswani2017attention}. The visual features are finally aggregated according to the learned relations. This fusion module utilizes the extra features more effectively; thus, the performances are further improved.

Most previous works \cite{tang2013content, Lu2015DeepMA, Ma2017ALampAL} adopted the classification task to predict a binary aesthetic label. However, as discussed by previous works \cite{cui2018distribution, jin2018predicting}, binary labels cannot reflect the subjectivity and diversity of human assessment. Some works proposed to predict the average score instead. The methods include direct regression \cite{kao2016hierarchical,hosu2019effective} and pairwise comparison \cite{lee2019image}, but these methods still cannot reflect the diversity of human assessment. Considering all these drawbacks, we decided to predict the aesthetic score distributions as in \cite{cui2018distribution, jin2018predicting, talebi2018nima, zhang2019gated}.

The contributions of this paper are summarized as follows:

\begin{itemize}
    \item We develop a novel method called PRP by applying RoM pooling in networks that take padded images as inputs. This method enables us to utilize padding to maintain a unified image size while eliminating its side effects. This is the first method in AQA that supports end-to-end batch training on arbitrary full-resolution images. 
    \item An SA module is proposed to remedy the shape information loss in RoM pooling layer. The original image aspect ratios are encoded and fused with visual features to make the model adapt to different image shapes.
    \item We find that it is inaccurate to evaluate images in the AVA dataset without theme information because of the existence of theme criterion bias. A theme-aware module is proposed to tackle this issue. The themes are encoded and combined with visual features to predict the aesthetic quality more accurately.
    \item We propose an attention-based fusion module to fuse extra features including shape features and theme features with image features. This module mines the relations between different features effectively. The experiments prove the effectiveness of the proposed method.
\end{itemize}

The remainder of this paper is organized as follows. We summarize the AQA related works in Section~\ref{section2} and provide the details of the proposed method in Section~\ref{section3}. The experiments are described in Section~\ref{section4}. Finally, we conclude our paper and analyze future works in Section~\ref{section5}.

\section{RELATED WORK}\label{section2}
\subsection{Photo Aesthetic Quality Assessment}
Traditional aesthetic quality assessment is based on handcrafted features and shallow classifiers \cite{freund1997decision,he2009robust}. Ke \textit{et al.} \cite{ke2006design} analyzed several aesthetic factors and connected them with low-level visual features, such as the edge distribution to reflect the simplicity of a photo. Luo \textit{et al.} \cite{luo2008photo} proposed to focus on the image foreground and extracting different features to describe it. Tang \textit{et al.} \cite{tang2013content} further stated that the assessment should be based on the content. For example, they designed features, especially for human photos. Many other works \cite{bhattacharya2010framework,dhar2011high,su2011scenic,li2010towards} are also based on these low-level handcrafted features and may emphasize different features in different situations. Some works \cite{bhattacharya2010framework} also introduced applications based on aesthetic quality assessment such as aesthetic enhancement. In addition to these aesthetic-related features, generic features such as the GIST \cite{oliva2001modeling} and SIFT \cite{lowe2004distinctive} were employed, and good performance was obtained \cite{marchesotti2011assessing}. Although many of these features are carefully designed, their representation power is still limited. 

In recent years, DNNs have become widely adopted models in many research areas, and many new techniques \cite{song2019geometry,fu2021high,ma2021contrastive} based on DNNs have widened the applications of automatic AQA. Lu \textit{et al.} \cite{lu2014rapid, lu2015rating} were the first to assess photo aesthetic quality based on deep neural networks. The authors designed a two-column architecture to learn features on both the global and local views. Kao \textit{et al.} \cite{kao2015visual} proposed to use a regression model instead of a classification model because a continuous score can deliver more precise information about aesthetics. Dong \textit{et al.} \cite{dong2015photo} used features from an ImageNet pretrained network to predict aesthetic binary labels. Tian \textit{et al.} \cite{tian2015query} designed a query-based model; they trained a network for each query image based on a subset of images that is strongly related to the query. Additionally, Kao \textit{et al.} \cite{kao2017deep} proposed a multitask learning model to predict aesthetic labels and semantic labels simultaneously. The authors explained that AQA is strongly associated with image semantics.

\begin{figure*} 
    \centering
    \includegraphics[width=0.95\linewidth]{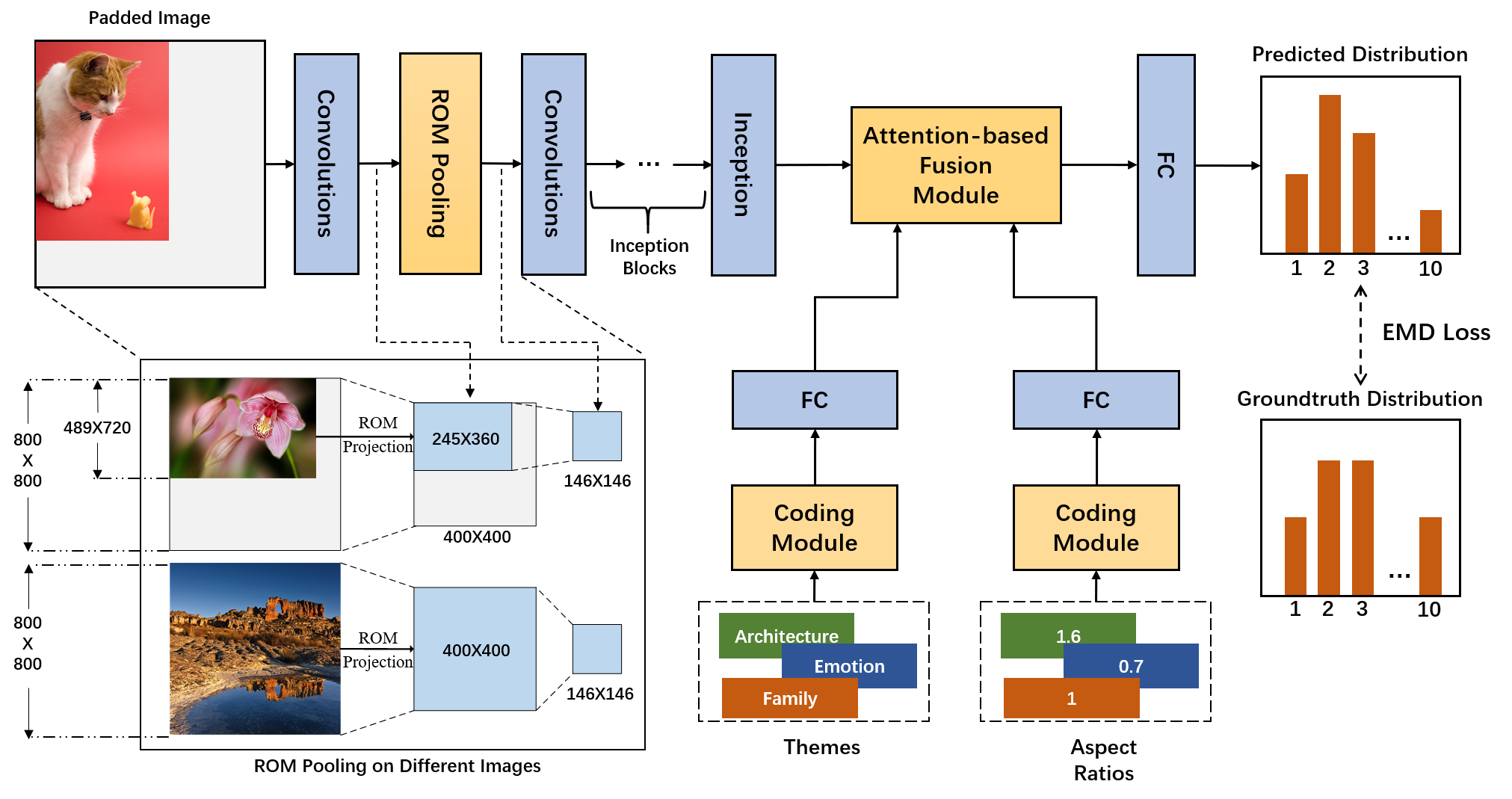}
    \caption{Overall architecture. The bottom-left region gives examples to show how images of different sizes are processed. The padded images are fed into the network. In RoM pooling, the features inside the RoM are pooled to a uniform size of $ 146 \times 146 $. The pooled features are then processed by several convolutional blocks and transformed into features of the shape $2048 \times 17 \times 17$. The themes and aspect ratios are encoded into one-hot codes in their coding modules and are turned into 2048-dimensional features by the corresponding fully connected layers. Finally, the visual features, the theme features, and the shape features are fused in the attention-based fusion module. The fused features are used to predict aesthetic distributions. EMD is employed as our loss function.}
  \label{frame} 
\end{figure*}

Although these works based on DNNs have made great progress, they still suffer from the fixed input size problem. Thus, researchers have proposed methods to solve this problem. Some works \cite{Lu2015DeepMA,Ma2017ALampAL,lu2015rating, wang2019aspect} proposed to crop multiple fixed-size patches from the original images, and aggregating the features extracted from these patches to predict the aesthetic quality. Ma \textit{et al.} \cite{Ma2017ALampAL} proposed to select patches according to some criteria based on human perception; therefore, the patches can be more representative. Instead of selecting patches according to predefined criteria, Sheng \textit{et al.} \cite{sheng2018attention} proposed an attention-based method that dynamically learns weights for different patches. Other works sought to maintain the aspect ratio of input images; they adopted either an FCN \cite{long2015fully} or SPP \cite{he2015spatial} to generate fixed-size features from arbitrary network inputs \cite{apostolidis2019image,fang2018image,Mai2016CompositionPreservingDP,cui2018distribution}. To adapt to the deep learning tools that are preferably run on fixed inputs, some used multisize training as an approximation, but only $ 1.0 $ and $ 1.5 $ aspect ratios were taken into account \cite{fang2018image, cui2018distribution}. Apostolidis \textit{et al.} \cite{apostolidis2019image} conducted experiments with a batch size of 1 so that any input size could be accepted. However, the performance of this method is even worse than that of the image transformation methods, possibly due to the unstable training process. Hosu \textit{et al.} \cite{hosu2019effective} extracted features from ImageNet pretrained models without fine-tuning; thus, they could use the original photos in a two-stage training process. Chen \textit{et al.}\cite{chen2020adaptive} proposed a different thought that uses convolutions of fractional strides to adapt to warped inputs.

In addition to various works on solving the fixed-size input problem, some other researchers have focused on different aesthetic evaluation forms. Most previous works used binary labels. Recently, other frameworks, including the aesthetic score regression, pairwise comparison, and distribution learning, were developed. Kao \textit{et al.} \cite{kao2015visual} used a regression model to evaluate images. Lee \textit{et al.} \cite{lee2019image} designed a unified framework based on pairwise comparison to achieve classification, regression and, personalization simultaneously. Similarly, Li \textit{et al.} \cite{li2020personality} also achieved both generic and personalized AQA via a multitask learning and fusion framework. The aesthetic distribution can reflect the diversity and subjectivity of image qualities. Therefore, it attracted the attention of many researchers. Some early works \cite{wu2011learning} have employed label distribution learning based on a support vector machine. They also used the voting number to represent the reliability of the ground-truth distribution. Recently, many other works predicting aesthetic rating distributions were proposed and there were various loss functions such as the Kullback-
Leibler (KL) divergence \cite{cui2018distribution}, earth mover distance \cite{talebi2018nima}, $ \chi^2 $ distance \cite{jin2016image}, and cumulative Jensen-Shannon divergence \cite{jin2018predicting}. These works also combined other strategies, such as using semantic information or defining reliability based on the distribution kurtosis.

\subsection{Pooling Methods and Multisize Feature Extraction Methods}
Pooling is one of the basic components in current deep neural networks. It plays a crucial role in reducing computational complexity. In the original implementations of pooling methods, the stride and the kernel size are manually fixed so that the ratio between the pooling input size and the output size is fixed. He \textit{et al.} \cite{he2015spatial} stated out that such a property makes the networks inflexible because the fully connected layers can only accept fixed-size inputs. SPP was designed to enable the pooling layers to generate fixed-size features from arbitrary inputs, and it can also extract multiscale information. This is achieved by using adaptive pooling kernels and strides. Specifically, the pooling size is manually fixed, and the pooling stride, as well as the kernel size, are defined as the ratio between the input size and the fixed pooling size. To incorporate SPP into networks, two types of methods were proposed. One uses a multisize end-to-end training strategy with two predefined sizes 224 and 180. The other uses a two stage training framework without fine-tuning the backbone CNN such that images do not need to be resized.

In both traditional pooling and SPP, pooling is applied to the entire input feature map. In other words, the moving range of the pooling kernel is fixed as the entire input. However, in the object detection task, an object only occupies a part of the image region. To extract features from only object regions and discard other features, RoI pooling was proposed in \cite{girshick2015fast}. RoI pooling is able to extract fixed-size features from regions of arbitrary sizes and locations. This means that the kernel moving range, the kernel size and the stride can be adaptively modified according to different RoIs. RoI pooling was improved to RoI align in \cite{he2017mask} to rectify the quantification error. This enhancement significantly improves the detection performance of small objects. RoI pooling (align) has become a standard module in many object detection and segmentation models \cite{dai2016r,lin2017feature,cai2018cascade}.

SPP and RoI pooling both enable the network to accept input images of arbitrary sizes. To achieve this goal, removing the fully connected layers and using only convolutional layers is also a solution. Such a network named an FCN was first introduced to solve semantic segmentation tasks \cite{long2015fully}.

\begin{figure} 
\centering

\begin{minipage}[b]{0.45\textwidth}
    \centering
  \subfloat[Crossing the line\protect\footnotemark \label{10a}]{%
       \includegraphics[height=4.8cm]{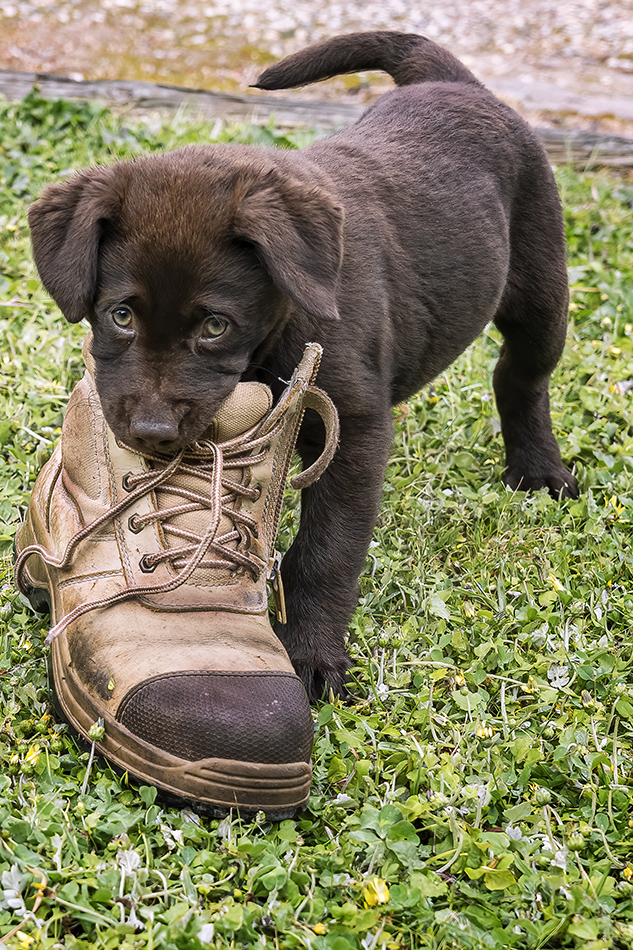}}
    \hfill
  \subfloat[Brown\protect\footnotemark\label{10b}]{%
        \includegraphics[height=4.8cm]{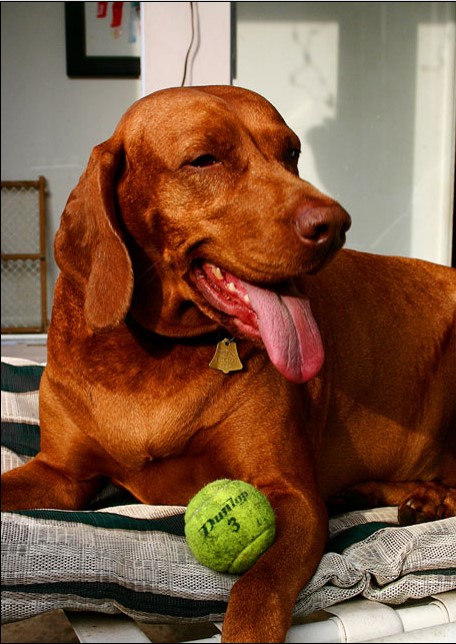}}
\end{minipage}

  \caption[Caption for LOF]{Examples of the images with the same semantic label but different themes. Both images are downloaded from DPChallenge.com.}
  \label{themesemantic}
  
\end{figure}
\footnotetext[1]{$\text{www.dpchallenge.com/image.php?IMAGE\_ID=1254363}$}
\footnotetext[2]{$\text{www.dpchallenge.com/image.php?IMAGE\_ID=439173}$}

\section{METHODS}\label{section3}
This section provides a detailed introduction of the proposed method. We first describe how to combine RoM pooling and image padding in Section~\ref{section3a} and introduce the shape-aware module in Section~\ref{section3b}. Then we introduce the theme-aware model in Section~\ref{section3c} and describe how to fuse shape features and theme features with visual features in Section~\ref{section3d}. Finally, we detail the training and inference procedure and network architecture in Sections~\ref{section3e}.

\subsection{RoM Pooling on Padded Full-Resolution Images}\label{section3a}

Image transformations in traditional DNNs cause mismatches between images and their annotations. Theoretically, SPP and an FCN can handle inputs of arbitrary sizes, but such an ability is not fully utilized because GPU implementations are preferably run on fixed inputs. Therefore, our method aims to remove the image-annotation mismatches while keeping the uniform input size to adapt to current deep learning tools. This goal is achieved by combining image padding with RoM pooling. In summary, padding turns the inputs into the same size, and RoM pooling eliminates the side effects that padding causes.

Specifically, we pad all images in datasets to the same size. It does not matter how the images are padded, so we use the most straightforward method that pads zeroes along the bottom and right boundaries. After padding, the input image becomes the spatially separable combination of two regions: one region is filled with padded values of 0, and the other region is the original image. The region of the original image is called the RoM, and its spatial coordinate range is from $(0, 0)$ to $(w, h)$, where $w$ and $h$ are the original image width and height, respectively.

The padded image is then fed into the network and processed by some convolutional layers. The RoM is mapped on the feature maps correspondingly. As a result, the input feature maps of the RoM pooling layer also become a spatially separable combination of a padded region and an image region (RoM). The RoM pooling layer pools features in the RoM to the manually defined size and the features outside the RoM are discarded. Therefore, the pooled features are of the same size, and the extra information introduced by padding is removed.

Formally, the feature at output location $(b,c,m,n)$ of the RoM pooling is given below.
\begin{equation}
{A}_{b,c,m,n}=\max\limits_{(i,j)\in\Omega_{mn}^b}\{f_{b,c,i,j}\},
\end{equation}
where $b$ is the batch index, $c$ is the channel index, $m,n$ are the spatial locations, $f$ is the RoM pooling input features, $A$ is the output pooled feature maps, and $\Omega_{mn}^b$ is the pooling kernel. The pooling stride is the same as the kernel size by default, and the kernel size is defined as the ratio of the pooling input RoM size to the manually defined pooling size $S_{pool}=(w_{out},h_{out})$. The pooling input RoM size is calculated as 
\begin{equation}
S_{RoM}^b=({\rm{round}}(\frac{w_b}{\tau}), {\rm{round}}(\frac{h_b}{\tau})),
\end{equation}
where $(w_b, h_b)$ is the original size of the $b^{th}$ image in one batch, $\tau$ is the downsampling ratio of the RoM pooling input size to the image padding size, and $\rm{round}(\cdot)$ is the rounding operation. Based on the pooling size $S_{pool}$ and the RoM size $S_{RoM}^b$, the kernel size is given as
\begin{equation}
S_{k}^b=({\rm{round}}(\frac{w_b}{{\tau}w_{out}}), {\rm{round}}(\frac{h_b}{{\tau}w_{out}})),
\end{equation}
The kernel moving range needs to be inside the RoM. Taking the top-left corner of the image to be $(0, 0)$, we give the exact definition of the RoM pooling kernel as follows.
\begin{equation}
\Omega_{mn}^b=[m{S_k^b(0)}, (m+1){S_k^b(0)}]\times[n{S_k^b(1)}, (n+1){S_k^b(1)}],
\end{equation}
Note that the kernel $\Omega_{mn}^b$ may be different for different input images.
The overall procedure is to apply the pooling operation to specific regions (RoM) that only contain features of the original images, as shown in Fig.~\ref{frame}. In the bottom left of the figure, light blue and gray regions denote the features from the RoM and padded region, respectively. The features inside the RoMs are eventually pooled into feature maps of the same size so that the network can be trained in an end-to-end manner.

From the RoM pooling process, we can easily observe the differences between RoM pooling, SPP, and traditional pooling. In traditional pooling methods, the kernel size and the pooling stride are manually assigned regardless of the inputs. Therefore, the pooling size cannot be kept unchanged for different inputs. In SPP, the pooling kernel scans the entire image; thus, we cannot discard the features in specific regions. Only in RoM pooling is the kernel moving range flexible and can the kernel size (pooling stride) be adaptively modified.

\emph{Discussion}: RoM pooling aims to achieve arbitrary pooling kernel sizes on a tensor; therefore, quantification approximations are introduced on both the region location and pooling stride. Such quantification error causes mismatches of RoMs. He \textit{et al.}~\cite{he2017mask} proposed to solve this problem by using interpolation; we call this improved version used in our model RoM align. RoM pooling and RoM align perform nearly the same in our model for two reasons. First, the mismatches of RoMs caused by quantification errors are determined by the downsampling ratio between network inputs and pooling inputs. For the downsampling ratio of $\tau$, the number of mismatched pixels along one dimension is at most $\lfloor{\frac{\tau}{2}}\rfloor$. Therefore, the small downsampling ratio (which will be discussed in Section~\ref{section3e}) brings small mismatches in our model. Second, the mismatches have fewer influences on larger regions. For example, mismatches of 10 pixels may totally change the location of a $10 \times 10$ region, but cause nearly no visible differences on a $500 \times 500$ region. Since the RoM is the entire full-resolution image in our model, the mismatches have nearly no impact. 

\subsection{Encoding Shape Information}\label{section3b}
It is widely known that image shapes influence human aesthetic perception. For example, images with some specific aspect ratios, such as 16:9, 4:3 and 1:1, are more popular in daily applications \cite{tu2020image, zeng2020grid}. Furthermore, images with extremely large or small aspect ratios may be unpopular. Revisiting our proposed PRP module, we find that although the model learns some knowledge from the original-shape images, the unified pooling output size of the RoM pooling layer still makes some shape information to be lost. This property will exert negative influences on our AQA model. To remedy this shortcoming, we propose a shape-aware (SA) module that encodes and extracts features from image aspect ratios and combines them with visual features. Specifically, we discretize the continuous aspect ratios, and turn the discrete values into one-hot codes. Then two fully connected layers are employed to extract shape features from the one-hot codes. Finally, the shape features are fused with visual features to predict the aesthetic quality. The SA module can be regarded as a patch to fill the loophole of the PRP module, making the model directly utilize shape information to improve the performance. We will introduce the feature fusion module in detail in Section~\ref{section3d}.

\begin{figure}[!t]
    \centering
    \includegraphics[width=0.8\linewidth]{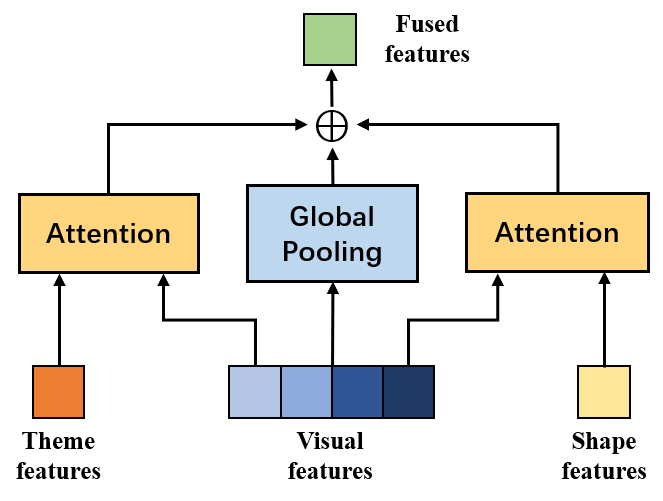}
    \caption{Attention-based fusion module. The module outputs are the summation of three features: (1) The outputs of an attention layer that takes theme features as query inputs and visual features as key and value inputs. (2) The outputs of an attention layer that takes shape features as query inputs and visual features as key and value inputs. (3) The global average pooling outputs of visual features.}
    \label{fusefig}
\end{figure}

\subsection{Theme-aware AQA}\label{section3c}
In the AVA dataset, images are submitted to the predefined challenges and assessed under specific challenge themes such as ``shapes'' and ``harsh environments''. Different aesthetic criteria are adopted in different themes. We call this phenomenon theme criterion bias. As illustrated in Fig.~\ref{themecriterion}, if people assess an image without its corresponding theme, the results may be inaccurate since the assessment criterion is improper.

Some previous works \cite{tang2013content, murray2012ava, cui2018distribution} proposed to assess images based on semantic labels. There are two major differences between semantic labels and themes. The first is the coverage difference. Semantic labels mainly describe objective contents such as ``human", ``landscape", ``city" and ``night" \cite{tang2013content}. The themes cover a wider range. Objects and abstract descriptions, such as ``balance", ``affluence" and ``second exposure" are included \cite{challenge_history}. The second difference distinguishes the two concepts from the view of the information sources. Themes are determined by humans subjective intents while semantic labels are determined by the images themselves and cannot be manually changed. We then use two phenomena to further demonstrate the differences. The first phenomenon is that images with the same semantic labels may belong to different themes. Fig.~\ref{themesemantic} shows an example in which the two images, both labeled with ``dog'', belong to different themes. The second phenomenon is that the themes are sometimes impossible to infer from images, but we can easily know the semantic labels. This is also obvious in Fig.~\ref{themesemantic}. It is nearly impossible to infer the theme ``crossing the line" for Fig.~\ref{themesemantic}(a) based on only the image itself.

From the previous analysis, we conclude that semantic labels are sometimes helpless in handling theme criterion bias. Since the themes cannot be obtained from the images themselves, they need to be provided along with the images in the evaluation process. To this end, we introduce a theme-aware (TA) module that fuses theme information with visual features. Specifically, we turn the 1,397 different themes in the AVA dataset into one-hot codes, and two fully connected layers are employed to extract theme features. The theme features are fused with the visual features using an attention-based fusion module. Finally, the fused features are fed into the head layers to predict the aesthetic qualities. The process is illustrated in Fig.~\ref{frame}. The detailed feature fusion process is introduced in Section~\ref{section3d}.

The theme features allow the network to adapt to different themes. This theme-adaptation advantage can be reflected in two aspects. First, visual aesthetic features from different themes are used differently. For example, suppose that two identical images may obtain different aesthetic quality assessments just because of their different themes. In this situation, only theme information can help the model to predict different aesthetic qualities to match the ground truth. Second, different images from the same themes are treated equally. Only visual features can influence the model assessments when the images share the same theme.

\begin{figure}[!t]
    \centering
    \includegraphics[width=0.77\linewidth]{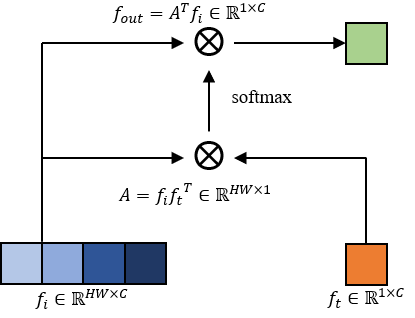}
    \caption{Attention module. First, we calculate the inner-product of the flattened visual features $f_i$ and the extra features $f_t$. The obtained attention matrix $A$ that reflects pairwise relations between visual features and extra features is then normalized with the softmax function. Finally, the visual features are aggregated based on the attention matrix $A$. Note that we do not plot the linear transformations applied to the input features, the normalization layers, and the activation function for simplicity.}
    \label{attentionfig}
\end{figure}

\subsection{Attention-Based Feature Fusion}\label{section3d}
In addition to the visual features, two types of extra features are used in our model. The shape features aim to remedy the lost shape information in the PRP module. The theme features introduce theme criterion bias such that the model learns aesthetic criteria more accurately. To effectively fuse the extra features with visual features, we propose an attention-based feature fusion module. The overall architecture of the module is displayed in Fig.~\ref{fusefig}. Two independent attention layers are used to process shape features and theme features, respectively. We also use a global average pooling layer to pool the visual features directly. As a result, we finally obtain three features, including shape-guided visual features, theme-guided visual features, and pure visual features. The three features are added together as the fusion outputs.

The two attention layers in Fig.~\ref{fusefig} are the same. In general, an attention process is formulated as follows \cite{vaswani2017attention},
\begin{equation}\label{attentioneq}
\text { Attention }(Q, K, V)=\operatorname{SoftMax}\left(Q K^{T} / \sqrt{d}+B\right) V,
\end{equation}
where $Q$, $K$, $V$ are query, key and value features respectively. $d$ is the feature dimension, and $B$ is the positional embedding. In our model, the extra features are regarded as queries $Q$ and the flattened visual features are used as keys $K$ and values $V$. The architecture of the attention layer is illustrated in Fig.~\ref{attentionfig}, where $f_t$ and $f_i$ are the extra features and the flattened visual features, respectively. In our implementation, we directly use the multihead attention proposed in \cite{vaswani2017attention} and set the number of heads to eight. Because the length of the query features $f_t$ is one, there is no need for positional embedding. The attention layer turns the spatial dimension of visual features to 1 and can be regarded as a special global pooling layer that assigns different weights to different spatial locations according to the extra features. This module mines the relations between extra features and image spatial features. Therefore, the extra features are utilized more effectively.

\begin{figure}[!t]
    \centering
    \includegraphics[width=0.89\linewidth]{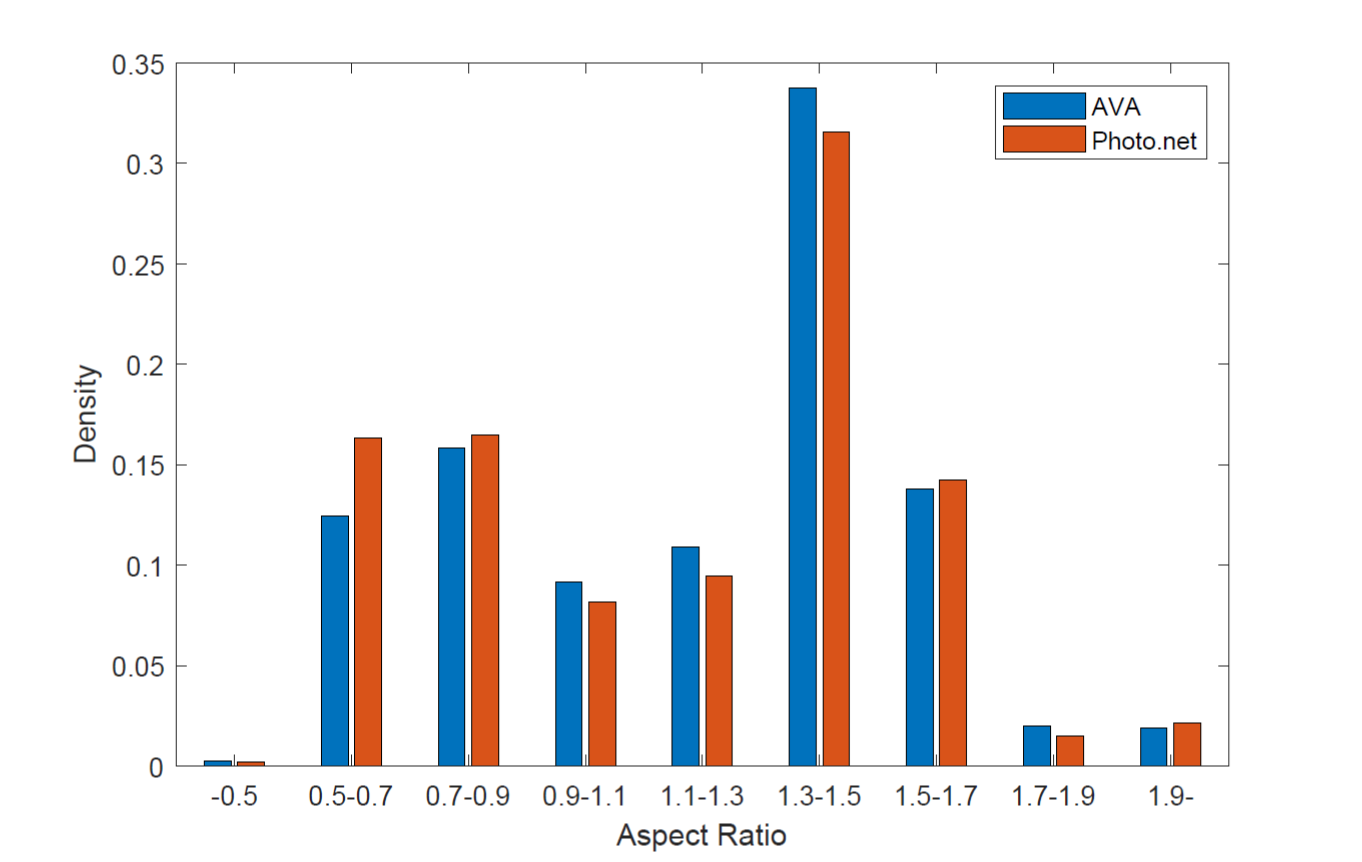}
    \caption{Distributions of image aspect ratios. Blue and red bins denote the distributions of the AVA and Photo.net datasets, respectively. We can see that the range is wide, showing the diversity of the aspect ratios. The figure also shows that the two datasets show nearly the same aspect ratio distribution.}
    \label{aspectratio}
\end{figure}

\begin{figure*} 
    \centering
    \includegraphics[width=0.99\linewidth]{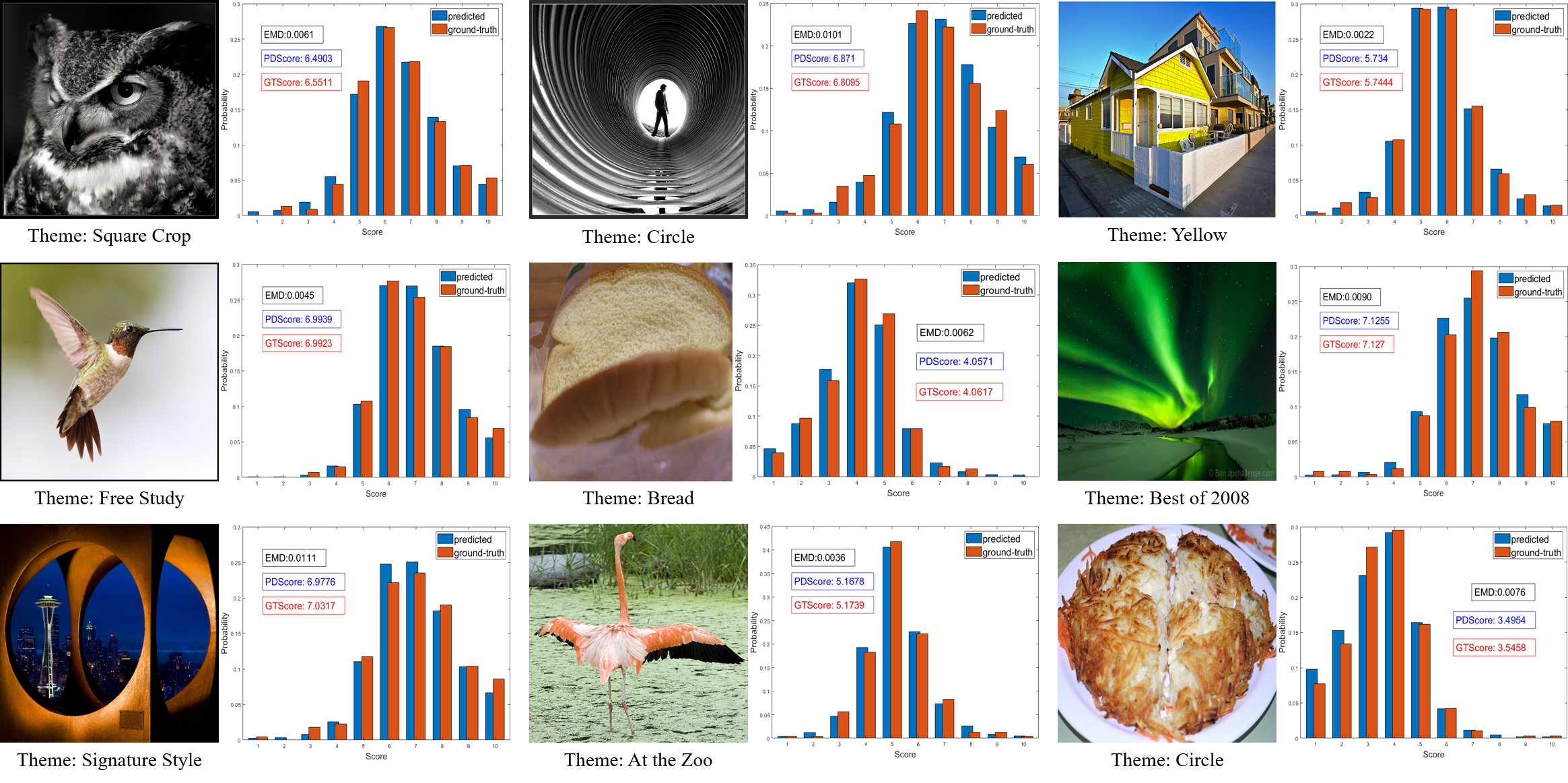}
    \caption{Some distribution prediction results. The blue bins are the predictions, and the red bins are the ground truth. The EMD between them and the mean scores (both ground-truth score (GTScore) and predicted score (PDScore)) are annotated in the corresponding figures. The corresponding themes are given below the images. The figure shows that our model adequately predicts distributions of both good (GTScore $>5$) and bad (GTScore $<5$) photos.}
    \label{distributionresults}
\end{figure*}

\subsection{Network Architecture and Learning Process}\label{section3e}

We choose Inception-V3 \cite{szegedy2016rethinking} as our backbone network, which is the same as previous works \cite{talebi2018nima, zhang2019gated, hosu2019effective}. We insert the RoM pooling layer in Inception-V3 to replace the original first pooling layer. Our model takes the padded images as the inputs. This is different from the original Inception-V3, which resizes all the inputs to $299\times299$. As a result, the RoM pooling size needs to be modified correspondingly. The RoM pooling size is a hyperparameter and there is no simple method to find an optimal solution. Therefore, we choose the pooling size based on some empirical clues. Our main consideration is to keep the ratio between the first pooling size and the original image size similar to that of the original Inception-V3. However, the size may be different for different images. Therefore, we choose a relatively common image size $600$ in the AVA dataset as the benchmark. As a result, the RoM pooling size $(w_{out},h_{out})$ is obtained by multiplying the benchmark size of $600$ with the corresponding ratio $\frac{73}{299}$ in the original Inception-V3
\begin{equation}
w_{out}=h_{out}=600\times\frac{73}{299}\approx146,
\end{equation}
where $73$ is the pooling size of the first pooling layer in the original Inception-V3.

In the training stage, we calculate the ground-truth aesthetic distribution $\bm{p}^i$ for the training image $\bm{x}^i$ by normalizing the numbers of votes on all the predefined quality scores. 
\begin{equation}
p^i_k=\frac{v^i_k}{\sum_{k=1}^K v^i_k},
\end{equation}
where $v^i_k$ is the number of votes for the quality score $k$, and $K$ is the maximum quality score. After obtaining the predicted aesthetic distribution $\widehat{\bm{p}}^i$ from our model, we employ the earth mover distance (EMD) \cite{talebi2018nima} between distributions as the loss function. The EMD loss is based on the cumulative distribution density, which works well when the distribution is ordered, as discussed in \cite{wu2011learning}. It is defined as
\begin{equation}\label{emdeq}
{\rm EMD}\bm{(\bm{p}^i,\widehat{\bm{p}}^i)}=(\frac{1}{K}\sum_{k=1}^{K}|{\rm CDF}_{\bm{p}^i}(k)-{\rm CDF}_{\widehat{\bm{p}}^i}(k)|^{r})^{\frac{1}{r}},
\end{equation}
where $ {\rm CDF}_{\bm p}(k) $ is the cumulative distribution of $ \bm{p} $ defined as $ \sum_{j=1}^{k}p_j $. We also choose $ r=2 $ for its simplicity in optimization.

In the testing stage, we adopt the same procedure as in the training stage. Testing images are padded to the same size and fed into the network. One may argue that using the original images without padding and removing the RoM pooling layer is better, but such a method leads to weak accordance between the testing and training stages. The learned network may not fit such a situation. We test this method and obtain worse results.

\begin{figure*} 
    \centering
    \includegraphics[width=0.99\linewidth]{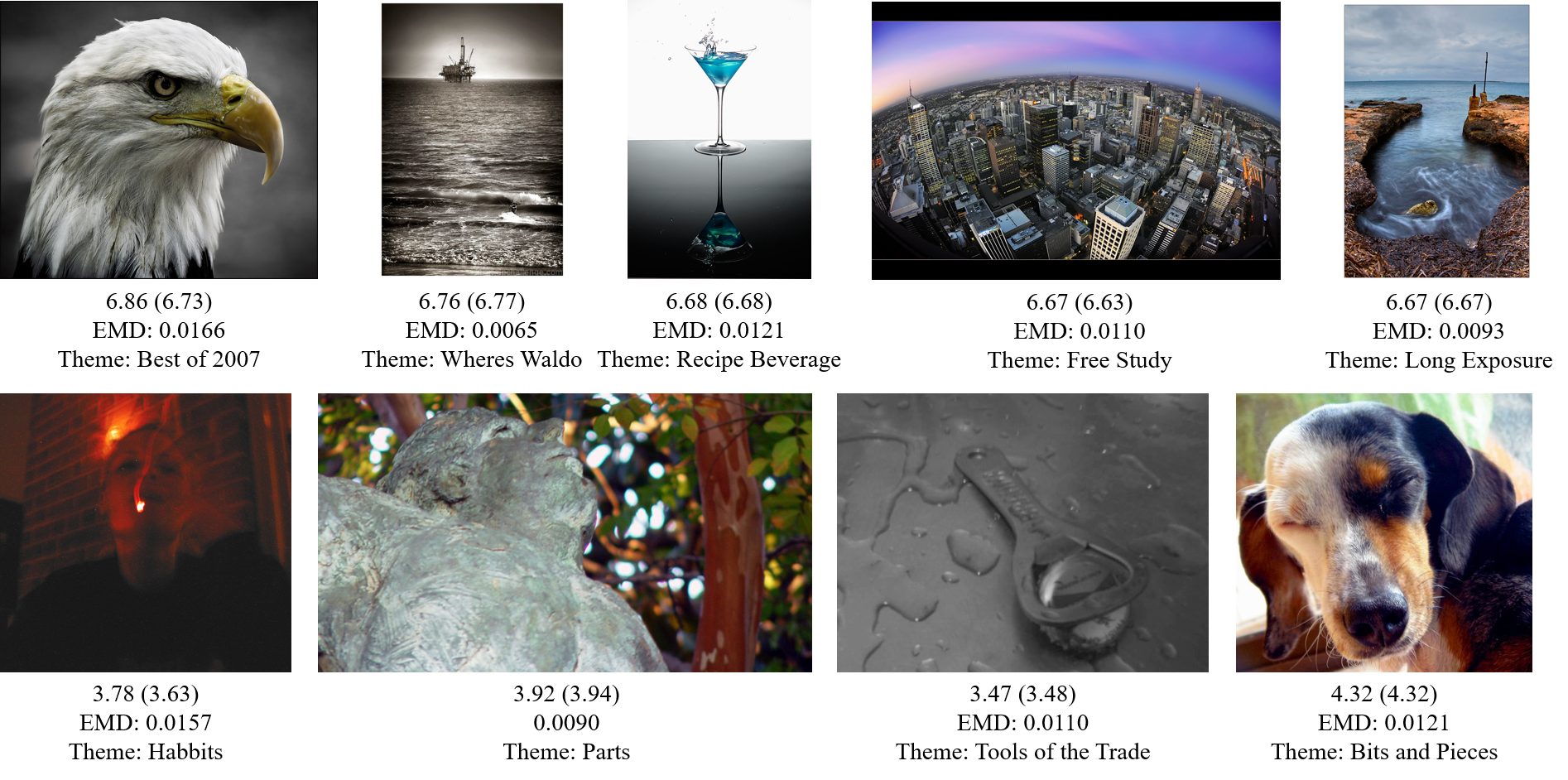}
    \caption{Some well-predicted images. The images are shown in their original aspect ratio. The predicted mean score (ground-truth score), EMD (r=1) and the corresponding themes are given below each image.}
    \label{scoreresults}
\end{figure*}

\section{EXPERIMENTS}\label{section4}

\subsection{Datasets and Evaluation Metrics}\label{section4a} 
We evaluate and compare our algorithm with other AQA algorithms on two widely used datasets, AVA \cite{murray2012ava} and Photo.net \cite{datta2008algorithmic}. Both datasets are collected from websites. The aspect ratio distribution is shown in Fig.~\ref{aspectratio}. We can see that both datasets contain various aspect ratios and cover a wide range. Surprisingly, although the two datasets come from different websites and have different scales, their aspect ratio distributions are nearly the same.
\subsubsection{AVA}
AVA is a large-scale database for image aesthetic quality analysis and contains over 250,000 color images. All images are collected from \url{www.DPChallenge.com}. The aesthetic assessment is given by 78 $\sim$ 549 individuals, and each of the voters chooses a score from 1 to 10. Each image belongs to one theme and there are 1,397 themes. The theme labels are provided in the dataset from the file ``Challenges.txt''. The dataset also provides other two types of annotations. The first is photographic style annotation, which is a sampled and merged subset of different themes and only covers a small part of the images. The second is semantic annotation. There are 66 different labels and each image contains at most two labels. As explained in Section~\ref{section3c}, semantic annotation and themes are different concepts. We follow the standard dataset partition as in \cite{murray2012ava}. There are approximately 19,817 images for testing. For the other images, we use 230,000 images for training and the remaining 4404 images for validation. The aesthetic annotations, semantic labels, and challenge themes can be found at \url{https://github.com/mtobeiyf/ava_downloader/}.
\subsubsection{Photo.net}
The Photo.net dataset only provides aesthetic labels. It contains 20,278 images and each image is rated by at least ten users using a score of 1 to 7. Only the mean score and standard deviation are given in some images because their voting information is lost. Some images have been lost due to several website updates; therefore, only 16,666 images can be downloaded. We randomly select 14,800 images as the training set, 1200 images as the testing set, and 666 images as the validation set. The dataset can be found at \url{https://ritendra.weebly.com/aesthetics-datasets.html}

We evaluate our method using three types of metrics. The first is for distribution prediction. We employ several distribution distance metrics, including the Euclidean distance (Euc), Kullback-Leibler (KL) divergence, Jensen-Shannon (JS) divergence, chi-square ($ \chi ^ 2 $) distance, EMD with $ r=1 $ in Eq.~\ref{emdeq}, and cosine distance (CD), according to previous works \cite{jin2018predicting, fang2018image, cui2018distribution, talebi2018nima, zhang2019gated}. All these metrics indicate better performances if the value is smaller. The second is based on the expectation (mean score) and standard deviation of the aesthetic distribution. For the predicted and ground-truth score distributions, we calculate their mean score and the standard deviation, respectively, and use correlation coefficients and the mean squared error (MSE) to evaluate the performances. The correlation coefficients include the Pearson linear correlation coefficient (PLCC) and Spearman rank-order correlation coefficient (SRCC). Larger coefficients indicate better performance. The MSE is only applied to the mean score due to the lack of previously reported results. The third type is for binary classification. The mean score of the distribution is binarized to obtain the class labels. We use classification accuracy (Acc) as the metric to evaluate our model.

\subsection{Implementation Details}

All our models are pretrained on ImageNet to accelerate the convergence and fine-tuned on the corresponding data sets. An SGD optimizer with momentum of 0.9 and weight decay of 0.0001 is used to train the network. The learning rate is divided by 2 every 10 epochs, and the model is trained for 40 epochs. We set the initial learning rate as $ 4\times10^{-3} $ on the convolutional layers; and for randomly initialized layers, the learning rate is 10 times larger. We test two types of padding implementations and get nearly the same performances. One pads all images to $800\times800$, the other pads batched images to their biggest size in this batch. The batch size is set as 64. Due to the memory limitation, each GPU can only process our padded images with a maximum batch size of 16. Therefore, we use 4 GPUs to train the model. In case the batch normalization layers learn the statistical information inaccurately under such a small batch size on a single GPU, we use the cross-GPU synchronized batch normalization. Data augmentation is used to ease the overfitting problem. Specifically, we randomly use cropping and horizontal flipping to augment the images. Four types of cropping augmentations are applied, and each crops the height and width of the original image from one of the four corners by $\frac{7}{8}$. In the test stage, we average the results on all the augmented images as the final predictions.

\begin{table}[!t]
\renewcommand{\arraystretch}{1.3}
\setlength{\tabcolsep}{1.7mm}
\caption{Performances of the distribution prediction on AVA. "Align" indicates that we replace RoM pooling with RoM align}
\label{table1}
\centering
\begin{tabular}{lcccccc}
\hline
Models &Euc&KL&JS&$ \chi ^ 2 $&EMD&CD\\
&&&&&(r=1)&\\
\hline
Talebi \textit{et al.} \cite{talebi2018nima}&-&-&-&-&0.050&-\\
Zhang \textit{et al.} \cite{zhang2019gated} &-&-&-&-&0.045&-\\
Wang \textit{et al.} \cite{wang2019aspect} &-&-&-&-&0.065&-\\
Li \textit{et al.} \cite{li2020personality} &-&-&-&-&0.047&-\\
Fang \textit{et al.} \cite{fang2018image}&0.144&0.120&-&-&-&0.056\\
Cui \textit{et al.} \cite{cui2018distribution}&\textbf{0.127}&0.094&-&-&-&0.042\\
Jin \textit{et al.} \cite{jin2018predicting}&0.158&-&0.037&0.068&-&-\\
\hline
Ours&0.132&\textbf{0.085}&\textbf{0.021}&\textbf{0.039}&\textbf{0.039}&\textbf{0.039}\\
Ours (Align)&0.132&\textbf{0.085}&\textbf{0.021}&\textbf{0.039}&0.040&\textbf{0.040}\\
\hline
\end{tabular}
\end{table}

\begin{table}[!t]
\renewcommand{\arraystretch}{1.3}
\caption{Performances of the Mean Score and Standard Deviation Prediction on the AVA Dataset.}
\label{table2}
\centering
\setlength{\tabcolsep}{1.8mm}{
\begin{tabular}{lccccc}
\hline
\multirow{2}{*}{Models}&SRCC$\uparrow$&PLCC$\uparrow$&SRCC$\uparrow$&PLCC$\uparrow$&MSE$\downarrow$\\
       &(mean)&(mean)&(std.dev)&(std.dev)&(mean)\\
\hline
Kao \textit{et al.} \cite{kao2015visual} &-&-&-&-&0.4501\\
Jin \textit{et al.} \cite{jin2016image} &-&-&-&-&0.3373\\
Kao \textit{et al.} \cite{kao2016hierarchical} &-&0.5214&-&-&0.3988\\
Kong \textit{et al.} \cite{kong2016photo} &0.5581&-&-&-&-\\
Talebi \textit{et al.} \cite{talebi2018nima} & 0.6120&0.6360&0.2330&0.2180&-\\
Chen \textit{et al.} \cite{chen2020adaptive} & 0.6489&0.6711&-&-&0.2706\\
Meng \textit{et al.} \cite{meng2018mlans} & 0.6730&0.6860&-&-&-\\
Li \textit{et al.} \cite{li2020personality} & 0.6770&-&-&-&-\\
Wang \textit{et al.} \cite{wang2019aspect} &0.6868&0.6923&-&-&0.2764\\
Zhang \textit{et al.} \cite{zhang2019gated} & 0.6900&0.7042&-&-&-\\
Hosu \textit{et al.} \cite{hosu2019effective} &0.7450&0.7480&-&-&-\\
\hline
Ours&0.7736&\textbf{0.7753}&\textbf{0.7562}&\textbf{0.7512}&\textbf{0.2305}\\
Ours (Align)&\textbf{0.7737}&0.7749&0.7551&0.7498&0.2311\\
\hline
\end{tabular}}
\end{table}

\subsection{Performance Evaluation}

To compare with previous models comprehensively, we evaluate the performance of our distribution prediction model with three different types of metrics as introduced in Section~\ref{section4a}. In addition, some well-predicted and failed examples are shown and analyzed to better reflect the performance. 

We first compare our method with previous distribution prediction methods \cite{fang2018image, cui2018distribution, jin2018predicting, talebi2018nima, zhang2019gated, wang2019aspect,li2020personality}, where \cite{fang2018image, cui2018distribution, zhang2019gated, wang2019aspect} use fully convolutional networks or crop patches of fixed sizes from the original images for fixed input size problem while \cite{talebi2018nima,jin2018predicting} propose new loss functions for the distribution learning problem. The results in Table~\ref{table1} show that our method achieves the best performance in five of the six metrics and obtains the second-best results on the Euc metric, indicating the superiority over the competitors.


Then, we calculate the mean score and the standard deviation of the distributions, and compare their performances with previous methods. Among the competitors, GPF-CNN \cite{zhang2019gated}, NIMA \cite{talebi2018nima} and WCNN \cite{jin2016image} are distribution prediction models, while the rest competitors are regression models. Specifically, \cite{kao2016hierarchical,kong2016photo} introduce extra useful information such as scenes and attributes to guide the aesthetic learning. Some works \cite{chen2020adaptive,wang2019aspect,zhang2019gated,hosu2019effective} try to avoid changing the image aspect ratios with different modules, such as fractional dilated convolutions. Other problems, including personality \cite{li2020personality}, sample weighting \cite{jin2016image} and feature fusion \cite{meng2018mlans}, are also studied. Although these models are effective, the results in Table~\ref{table2} show that our method represents a new state-of-the-art approach. The performance improvement of 0.0287 on the SRCC (mean) is significant. Furthermore, our model obtains more than three times higher SRCC (std.dev) and PLCC (std.dev) than the previously reported best performances \cite{talebi2018nima}. These results effectively prove the merits of our model.

Finally, we calculate the mean scores of the aesthetic distributions and binarize them using a threshold of 5 to perform binary classification. Previous aesthetic quality classification methods mainly employ multi-patch aggregation \cite{Lu2015DeepMA,Ma2017ALampAL,sheng2018attention} methods or introduce semantic information \cite{kao2017deep}. The results in Table~\ref{accresults} show that our method achieves the third-best performance. Note that our model employs the EMD loss to learn aesthetic score distributions, while A-Lamp \cite{Ma2017ALampAL} and MP-Adam \cite{sheng2018attention} are binary classification models. Thus, our model can be applied to a wider variety of tasks.

\begin{figure}[!t]
    \centering
    \includegraphics[width=0.95\linewidth]{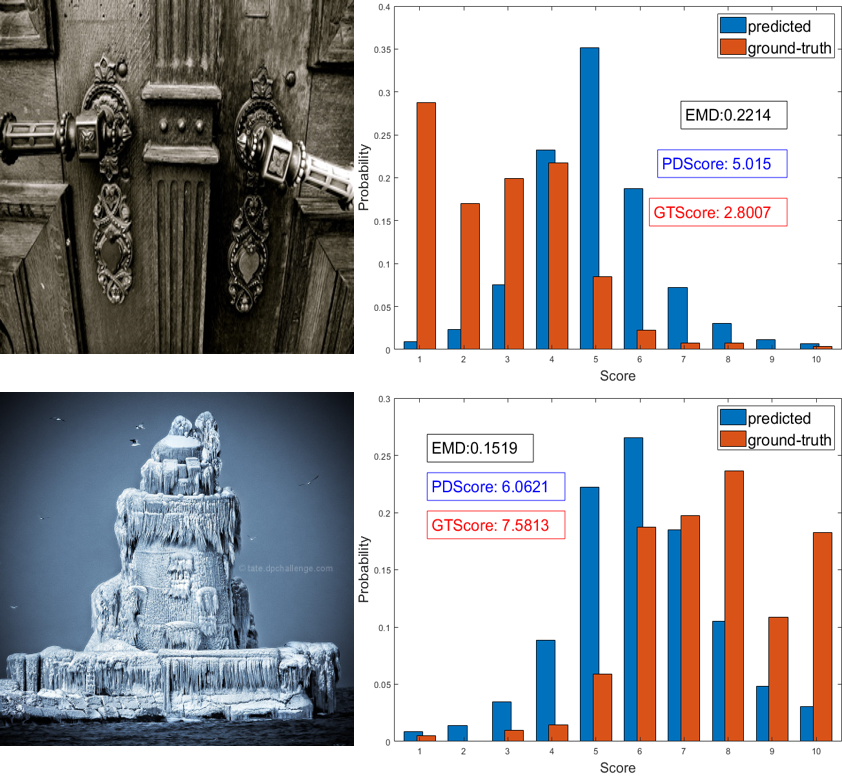}
    \caption{Failure cases. The model fails to predict uncommon distributions. The two distributions are both non-Gaussian and have abnormal values. In the first image, the number of people who vote for score 9 is too small. For the second image, many users vote for the score 1.}
    \label{failureimage}
\end{figure}

The results of the model that replaces RoM pooling with RoM align are also given. Both results are nearly the same, which proves our analysis in Section~\ref{section3a}. For simplicity, we only use RoM pooling in other experiments. We show some distribution prediction results on the AVA dataset in Fig.~\ref{distributionresults}. The blue bins represent prediction, and the red bins represent the ground truth. The results show that our method can precisely predict distributions. Other results are given in Fig.~\ref{scoreresults}, in which we keep the aspect ratio of the displayed images but only show the predicted (ground-truth) mean score and the EMD. Two failure cases are shown in Fig.~\ref{failureimage}. The figure shows that the ground-truth distributions of the two images are abnormal. They are both nontypical distributions in the AVA dataset, such as Gamma distribution \cite{murray2012ava}. This may be the reason why the model cannot assess them well.

\begin{table}[!t]
\centering
\renewcommand{\arraystretch}{1.3}
\caption{Binary Classification Performances on the AVA Dataset.}
\label{accresults}
\setlength{\tabcolsep}{1.8mm}{
\begin{tabular}{l|ccc}
\hline
Models &NMA-Net\cite{Lu2015DeepMA}&ADB-CNN \cite{kong2016photo}&MNA-CNN\cite{Mai2016CompositionPreservingDP}\\
\hline
Acc(\%) &75.4&77.3&77.4\\
\hline
\hline
Models&MTRL-CNN\cite{kao2017deep}&NIMA\cite{talebi2018nima}&POOL-3FC\cite{hosu2019effective}\\
\hline
Acc(\%) &79.1&81.5&81.7\\
\hline
\hline
Models &A-Lamp\cite{Ma2017ALampAL}&MP-Adam\cite{sheng2018attention}&Ours\\
\hline
Acc(\%) &82.5&\textbf{83.0}&82.4\\
\hline
\end{tabular}}
\end{table}

\subsection{Ablation Study}

To validate the effectiveness of each module, we use the methods based on three traditional transformations (the transformations shown in Fig.~\ref{traditionaltransform}) as the baseline models. The first baseline model resizes images to the same size. Since full-resolution images are used in our method, for a fair comparison, we resize all images to $800\times800$, which is the largest image size in the AVA dataset. The second baseline model uses padding transformation. We pad all images to $800\times800$. In both the resizing and padding baselines, to eliminate the influences of other factors, we use the same data augmentation methods and feature map size (the output size of the first pooling layer) as our full model experiments. The third baseline that randomly crops fixed-size patches needs to be combined with image resizing because the size of cropped patches cannot be larger than the smallest image size in the dataset. However, the smallest size in AVA is only $160\times160$. The input of this size cannot be used in Inception-V3 because the feature maps in the deep layers will be smaller than the convolutional kernels. Moreover, cropping such a small patch from much larger images will cause a very significant information loss, which leads to unfair comparisons. To conduct a feasible and convincing ablation study, we resize images to make the short edges no shorter than 512 while maintaining the aspect ratios. Then, we randomly crop $512\times512$ patches from all images to train the network. In the cropping baseline, only random flipping augmentation is needed. For simplicity, we choose the SRCC of the mean score and standard deviation, EMD with r=1 and KL divergence as the representative metrics. 

There are many factors to be validated. To demonstrate our contributions clearly, we split the experiments into three groups.

\subsubsection{PRP and SA Modules}
As described before, the proposed PRP module aims to replace the three traditional image preprocessing transformations, and the SA module is used to remedy the shape information loss in RoM pooling. Therefore, we compare them with the three baseline results to show their effectiveness. The results are given in Table~\ref{ablation1}. The table shows that the PRP module outperforms the three baselines. With the help of the SA module, the performances are further improved. For example, the performance gain of the SRCC (mean) metric is 0.0115 compared with the best baseline result, indicating the superiority of our method over the traditional transformations.

\begin{table}[!t]
\centering
\renewcommand{\arraystretch}{1.3}
\caption{Ablation Study I.}
\label{ablation1}
\begin{tabular}{cc|cccc}
\hline
\multicolumn{2}{c|}{\multirow{2}{*}{Models}}&SRCC$\uparrow$&SRCC$\uparrow$&\multirow{2}{*}{EMD$\downarrow$}&\multirow{2}{*}{KL$\downarrow$}\\
&&(mean)&(std.dev)&&\\
\hline
\multicolumn{1}{l|}{\multirow{3}{*}{Baselines}} 
&Resize&0.7353&0.3314&0.044&0.098\\
\multicolumn{1}{l|}{}&Pad&0.7328&0.3402&0.044&0.098\\
\multicolumn{1}{l|}{}&Crop&0.7287&0.3188&0.045&0.099\\
\hline
\multicolumn{2}{c|}{PRP}&0.7438&0.3424&0.043&0.097\\
\multicolumn{2}{c|}{PRP (SA)}&0.7469&0.3551&0.043&0.096\\
\hline
\multicolumn{2}{c|}{Resize + TA}&0.7644&0.7025&0.041&0.088\\
\multicolumn{2}{c|}{Pad + TA}&0.7633&0.7069&0.041&0.088\\
\multicolumn{2}{c|}{Crop + TA}&0.7615&0.6993&0.041&0.090\\
\multicolumn{2}{c|}{PRP (SA) + TA}&\textbf{0.7736}&\textbf{0.7562}&\textbf{0.039}&\textbf{0.085}\\
\hline
\end{tabular}
\end{table}

\subsubsection{TA Module}
Table~\ref{ablation1} gives the comparison results between nontheme models and TA models. The table shows that the performance improvements are consistent and significant. For example, after introducing theme information, the PRP (SA) model improvements the SRCC (mean) metric by 0.0268. Table~\ref{ablation1} also shows that compared with the SRCC (mean), the improvement on the SRCC (std.dev) is very substantial. This phenomenon proves that the theme-aware model can better reflect the subjectivity and diversity of the AQA task.

\begin{figure}[!t]
\centering
\includegraphics[width=0.99\linewidth]{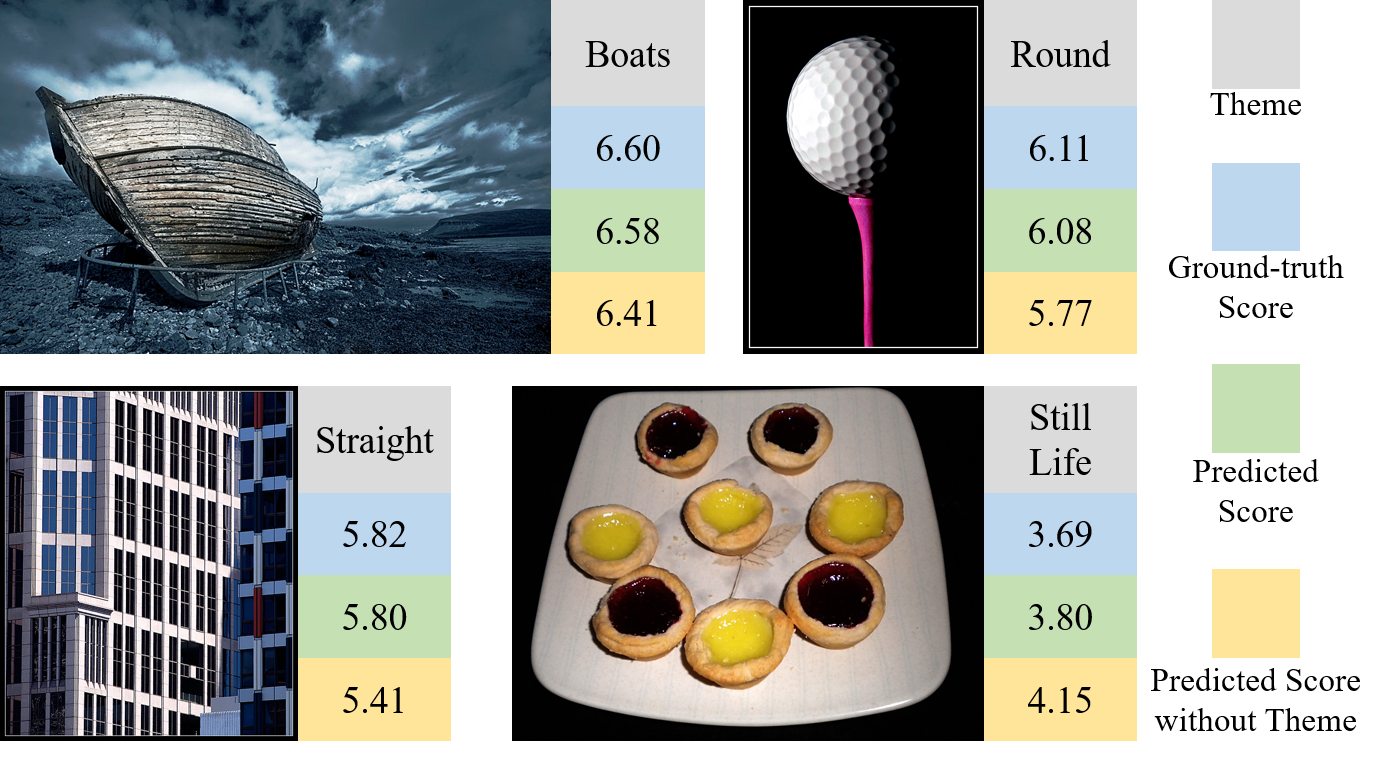}
\caption{Examples of the prediction results with and without theme information. We only showed the mean score. We can see that the model can predict a more accurate score with theme information.}
\label{withouttheme}
\end{figure}

To further show the influences of theme information, we give some AQA examples with two different theme conditions. In the first condition, the model predicts image aesthetic qualities without theme information. Some comprehensive examples are shown in Fig.~\ref{withouttheme}. The figure shows that for the first three images (images in the first row and the left image in the second row), the model tends to predict lower scores without theme information. One possible reason is that the model loses a positive aesthetic factor such that the images match the corresponding themes well. For example, the third image that contains many straight lines matches the theme ``straight'' well. 

\begin{table}[!t]
\centering
\renewcommand{\arraystretch}{1.3}
\caption{Ablation Study II.}
\label{ablation2}
\begin{tabular}{cc|cccc}
\hline
\multicolumn{2}{c|}{\multirow{2}{*}{Models}}&SRCC$\uparrow$&SRCC$\uparrow$&\multirow{2}{*}{EMD$\downarrow$}&\multirow{2}{*}{KL$\downarrow$}\\
       &&(mean)&(std.dev)&&\\
\hline
\multicolumn{2}{c|}{PRP}&0.7438&0.3424&0.043&0.097\\
\hline
\multicolumn{1}{c|}{\multirow{2}{*}{SA}} 
&Concatenate&0.7433&0.3468&0.043&0.097\\
\multicolumn{1}{c|}{}&Attention&0.7469&0.3551&0.043&0.096\\
\hline
\multicolumn{1}{c|}{\multirow{2}{*}{TA}} 
&Concatenate&0.7611&0.6918&0.041&0.088\\
\multicolumn{1}{c|}{}&Attention&0.7695&0.7358&0.040&0.086\\
\hline
\multicolumn{1}{c|}{\multirow{2}{*}{SA + TA}} 
&Concatenate&0.7620&0.7002&0.041&0.088\\
\multicolumn{1}{c|}{}&Attention&\textbf{0.7736}&\textbf{0.7562}&\textbf{0.039}&\textbf{0.085}\\
\hline
\end{tabular}
\end{table}

\begin{figure}[!t]
    \centering
    \includegraphics[width=0.95\linewidth]{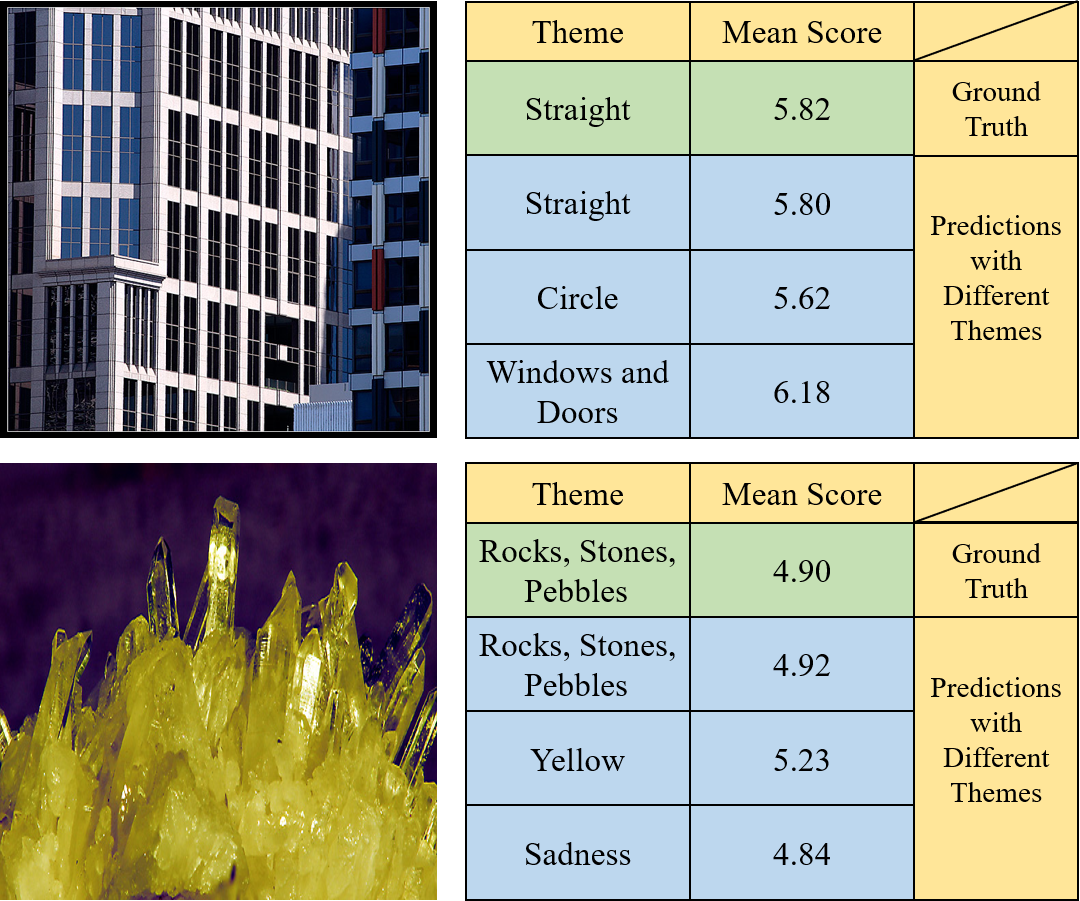}
    \caption{Examples of assessing aesthetic quality with different manually assigned themes. The assigned themes are selected from existing themes in the AVA dataset. The green boxes contain the default themes and the ground-truth average scores. Manually assigned themes and the corresponding predicted scores are displayed in the blue boxes.}
    \label{differenttheme}
\end{figure}

In the second condition, the model evaluates images under false themes, and two examples are shown in Fig.~\ref{differenttheme}. The first image that belongs to the theme ``straight'' contains a building with many rectangular windows. After replacing the input theme with ``circle'', the predicted score becomes lower. The possible reason is that there are no circles in the image; thus, it is very different from those high-score images in the theme ``circle''. If we use a more suitable theme of ``windows and doors'', a better result is obtained. The second image obtains a low average score. However, if it is evaluated with the theme ``yellow'', the predicted score becomes better. A possible reason is that the dominant color of the image is yellow. In contrast, using the theme ``sadness'' results in a lower score. The two examples show that different themes actually have different influences on AQA.

\subsubsection{Fusion Module}
We propose an attention-based fusion module to effectively utilize the extra features including shape features and theme features. To validate its effectiveness, we create a simple feature fusion module for comparison. This module directly concatenates the extra features with globally pooled visual features. The input dimension of the last fully connected layer is adjusted correspondingly. We compare the two different fusion modules using different extra feature combinations. From the results shown in Table~\ref{ablation2}, we can obtain three conclusions. First, the superiority of attention-based feature fusion is consistent on all three extra feature combinations, indicating the effectiveness of the proposed method. Second, although the performances are worse than those of attention-based fusion, the direct concatenation method is still helpful in TA models. Third, it is nearly useless to directly concatenate shape features with visual features.

\subsection{Theme Coding Method Analysis}

In our model, we encode the themes into one-hot codes. The question is whether there are some better choices to encode the themes. In this section, we discuss and validate some other coding methods.

\subsubsection{Binary Code}
In the binary coding method, we sort the themes in the AVA dataset in random order and turn the order numbers into binary system codes. The shortest code length is 11 because the number of themes 1,397 is larger than $2^{10}$ and smaller than $2^{11}$. As a result, this method reduces the theme code dimension from 1,397 to 11 and generates more compact theme codes.

\subsubsection{BERT Pretrained Pooled Phrase Embedding}
How to turn words, phrases, and sentences into meaningful embeddings has long been a hot research topic in the natural language processing (NLP) research community. In recent years, a milestone model called the bidirectional encoder representation from transformers (BERT) \cite{devlin2018bert} has attracted much attention. It has achieved promising results in many NLP downstream tasks. We use the pretrained model from Huggingface-Transformers \cite{wolf-etal-2020-transformers} package. The 768-dim feature before the BERT CLS-Head is used for each theme word. We pool the embeddings of all words in a theme to obtain the 768-dim theme embedding.

\begin{table}[!t]
\centering
\renewcommand{\arraystretch}{1.3}
\caption{Results of Different Theme Coding Methods.}
\label{themecode}
\begin{tabular}{c|c|cccc}
\hline
\multicolumn{2}{c|}{\multirow{2}{*}{Models}} & SRCC$\uparrow$& SRCC$\uparrow$ &\multirow{2}{*}{EMD$\downarrow$}&\multirow{2}{*}{KL$\downarrow$}    \\
\multicolumn{2}{c|}{} & (mean) & (std.dev) &       &       \\ \hline
\multicolumn{2}{c|}{PRP (SA)} & 0.7469 & 0.3551    & 0.043 & 0.096 \\ \hline
\multirow{3}{*}{\begin{tabular}[c]{@{}c@{}}PRP (SA)\\ +\\ TA\end{tabular}} & Binary  & 0.7422 & 0.3460    & 0.044 & 0.097 \\
& BERT    & 0.7563 & 0.6485    & 0.042 & 0.091 \\
& One-hot & \textbf{0.7736} & \textbf{0.7562}    & \textbf{0.039} & \textbf{0.085} \\ \hline
\end{tabular}
\end{table}

We test both types of new coding methods. Unfortunately, neither provides better results than the one-hot codes. The binary theme coding method even provides worse results than models that do not use theme information. The results are listed in Table~\ref{themecode}. 
Here, we provide some possible explanations for this phenomenon. The binary code eliminates the sparseness of the one-hot code; but introduces two extra weaknesses. First, the distances between different theme codes are reduced, which may influence the discriminability between them. Second, the distances between different theme codes are different, and this distance difference is randomly defined by the coding order, which may become harmful in some cases. The phrase embeddings from the pretrained BERT model are also more compact than one-hot codes. Furthermore, the method encodes some meaningful information. For example, the $L2$ distance between the theme codes of ``landscape'' and ``urban landscape'' is 2.29, which is much smaller than the distance of 4.79 between ``landscape'' and ``dog''. However, the model still obtains worse results when using such semantic embeddings. The worse performances may be the result of two factors. First, the BERT codes have lower discriminability between different themes. Second, our model cannot effectively utilize the semantic information in the BERT codes. We believe that the semantic information in BERT codes is valuable. We will further study this coding method in future works.

\begin{table}[!t]
\centering
\renewcommand{\arraystretch}{1.3}
\caption{Results on Different Pooling Sizes.}
\label{table4}
\begin{tabular}{l|llll}
\hline
Pooling&SRCC$\uparrow$&SRCC$\uparrow$&\multirow{2}{*}{EMD$\downarrow$}&\multirow{2}{*}{KL$\downarrow$}\\
Size &(mean)&(std.dev)& & \\
\hline
73&0.7609&0.6994&0.042&0.090\\
110&0.7688&0.7113&0.041&0.087\\
146&\textbf{0.7736}&\textbf{0.7562}&\textbf{0.039}&\textbf{0.085}\\
192&0.7625&0.7014&0.042&0.089\\
\hline
\end{tabular}
\end{table}

\subsection{Analysis of the RoM Pooling Size}

In this section, we analyze the influences of different RoM pooling sizes. Compared with the pooling size $73\times73$ of the first pooling layer in the original Inception-V3, the RoM pooling size in our model is increased to $146\times146$ to keep the pooling downsampling ratio similar to the original Inception-V3. To validate the importance of the pooling size, we conduct experiments to test the performances of different pooling sizes.

Theoretically, any variants of the pooling size may change the model performances. However, it is impractical to test all the possibilities. Therefore, we choose 4 representative pooling sizes, including 73, 110, 146, and 192 (the pooling width equals the pooling height). Note that the pooling size 146 is our default model setting. The results are shown in Table~\ref{table4}. We can clearly see that the performances improve when the pooling size is increased from 73 to 146. However, the improvement from 110 to 146 is not large, which means that the performances may be near saturation. Surprisingly, when the pooling size reaches 192, the performances become worse. One possible reason is that the network cannot comprehensively capture the global information from such large feature maps. Specifically, the receptive field of each layer is fixed. If the feature map sizes of the deep convolutional layers are too large, the long-range dependencies, which are important in the AQA task, may not be able to be effectively captured by the networks.

\begin{figure} 
    \centering
    \includegraphics[width=0.95\linewidth]{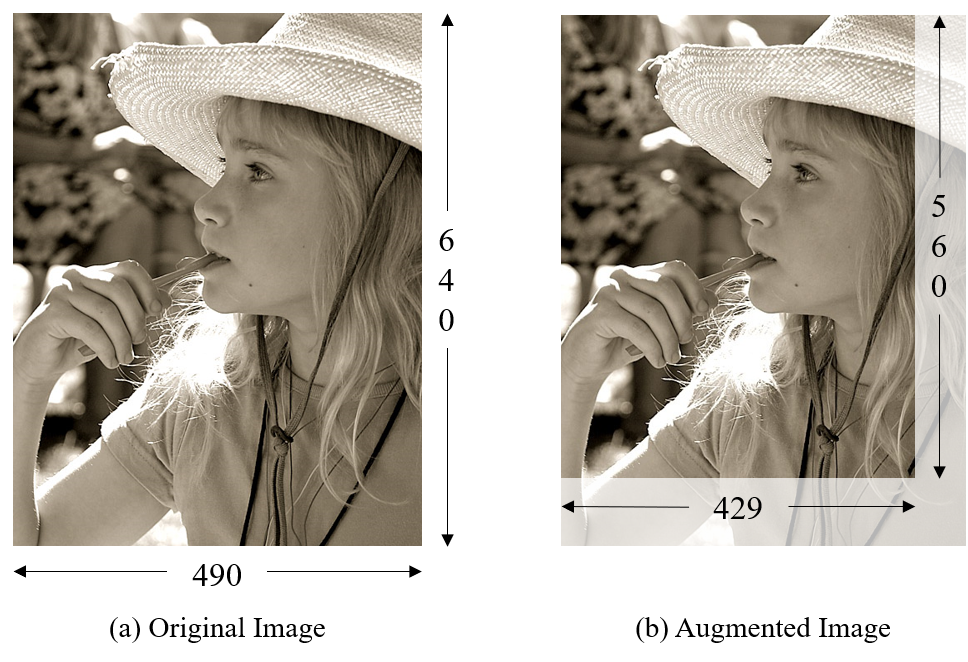}
    \caption{Examples of an image (a) and one of its cropping augmentations (b). The mean score of this image is 6.47. The aesthetic differences between the original and augmented images are slight.}
    \label{augmentation} 
\end{figure}

\begin{table}[!t]
\centering
\renewcommand{\arraystretch}{1.3}
\caption{Results on Different Data Augmentations.}
\label{table5}
\begin{tabular}{llllll}
\hline
\multirow{2}{*}{Augmentation}&SRCC$\uparrow$&SRCC$\uparrow$&\multirow{2}{*}{EMD$\downarrow$}&\multirow{2}{*}{KL$\downarrow$}\\
&(mean)&(std.dev)& & \\
\hline
Flip&0.7601&0.6994&0.0408&0.088\\
Flip+Crop (0.975)&0.7700&0.7331&0.0409&0.087\\
Flip+Crop (0.500)&0.7553&0.6973&0.0414&0.088\\
Flip+Crop (0.875)&\textbf{0.7736}&\textbf{0.7562}&\textbf{0.0393}&\textbf{0.085}\\
\hline
\end{tabular}
\end{table}

\subsection{Data Augmentation}

Data augmentation is a common process in deep learning frameworks. The traditional method includes random image cropping and flipping. As discussed before, cropping fixed-size patches from images may change the aesthetic information. Based on this consideration, some previous works in AQA \cite{cui2018distribution,fang2018image} only employed random flipping to augment images. However, one recent work \cite{hosu2019effective} used both image cropping and flipping in data augmentation and reported a performance improvement. The reason for the different results is the different degrees of cropping. Here, we call the operation that crops fixed-size patches as preprocessing cropping and call the operation in literature \cite{hosu2019effective} that aims to augment images augmentation cropping.

Specifically, in preprocessing cropping, because both the size and the aspect ratio differ considerably between images, the cropped patches from different images need to be small. Otherwise, it is impossible to keep the size of patches from different images the same. Small cropped patches result in significant information loss. However, in augmentation cropping, we do not need to keep the size of cropped patches the same. We augment images with proportional crops at $ \frac{1}{8} $ of the width and height. There is little damage to the aesthetic information with such a small discarded region near the edges. We give examples of cropping augmentation in Fig.~\ref{augmentation}. The figure shows that the differences between the cropped image and the original image are slight. In AQA, the assessment from one person always remains unchanged when there are only slight changes in the image. Therefore, the model can capture the consistency of AQA with cropping augmentation. To further validate our conclusion, we conduct experiments on different augmentation policies. As seen in Table~\ref{table5}, very small and large cropping ratios lead to poor performances.

\begin{table}[!t]
\centering
\renewcommand{\arraystretch}{1.3}
\caption{Performance on Photo.net dataset.}
\label{table6}
\begin{tabular}{llllll}
\hline
\multirow{2}{*}{Models}&SRCC$\uparrow$&PLCC$\uparrow$&\multirow{2}{*}{MSE$\downarrow$}&\multirow{2}{*}{EMD$\downarrow$}\\
       &(mean)&(mean)&&\\
\hline
Zhang \textit{et al.} \cite{zhang2019gated} &0.5217&0.5464&0.2715&0.070\\
Ours w/o CropAug&0.5687&0.5774&0.2231&0.066\\
AVA trained&0.5601&0.5790&0.2235&-\\
Ours&\textbf{0.5866}&\textbf{0.5903}&\textbf{0.2179}&\textbf{0.065}\\
\hline
\end{tabular}
\end{table}

\subsection{Evaluation on Photo.net}
Photo.net is a small dataset. Because there is no theme information, we only evaluate the model with RoM pooling on padded images. Analogous to the AVA dataset, we also pad images to $800\times800$. Personal subjectivity may have more significant impacts because the number of voters per image in this dataset is much smaller. This can result in unstable ground-truth distributions. Predicting distributions is harder in such a condition. To the best of our knowledge, there is only one previous work \cite{zhang2019gated} predicting aesthetic distributions on this dataset. They proposed to combine global and local image views to learn aesthetic distribution. Since they did not report the data augmentation strategy, we give results with and without cropping augmentation for a fair comparison. Table~\ref{table6} shows that our proposed method outperforms previous work. We also use this dataset to test the model trained on the AVA dataset to evaluate the generalization performance. Considering that the score ranges are different, the distribution metrics cannot be used. First, we obtain normalized scores between 0 to 1 from the AVA trained model. Then, the normalized scores are transformed to the range of 1 to 7. Finally, we compute the SRCC, PLCC, and MSE on the mean scores. The results are given in Table~\ref{table6}. We can see that the performances are still better than those of previous works.

\section{Conclusion and Future Works}\label{section5}

This paper proposes a simple but effective framework that supports end-to-end batch training on original full-resolution photos to predict their aesthetic distributions. We achieve this goal by combining image padding with RoM pooling. Padding turns the inputs to the same size, and RoM pooling removes features from the padded region. We also propose a shape-aware module that fuses aspect ratio information with visual features to remedy the shape information loss in RoM pooling. Furthermore, we found that evaluating photo aesthetic quality solely from images neglects theme criterion bias. Therefore, we use themes as extra information to learn aesthetic criteria precisely. Finally, an attention-based feature fusion module is proposed to effectively utilize shape information and theme information. The experimental results show that our method outperforms the state-of-the-art distribution learning and regression AQA models. Although our proposed method has achieved promising performances, additional work remains. For example, our method works poorly on skewed distributions. Therefore, studying how to predict such distributions more precisely is worth further study.


%




\ifCLASSOPTIONcaptionsoff
  \newpage
\fi



%

\small
\bibliographystyle{unsrt}
\bibliography{references.bib}

\begin{thebibliography}{10}

\bibitem{deng2018aesthetic}
Yubin Deng, Chen~Change Loy, and Xiaoou Tang.
\newblock Aesthetic-driven image enhancement by adversarial learning.
\newblock In {\em Proceedings of the 26th ACM International Conference on
  Multimedia}, pages 870--878. ACM, 2018.

\bibitem{ke2006design}
Yan Ke, Xiaoou Tang, and Feng Jing.
\newblock The design of high-level features for photo quality assessment.
\newblock In {\em Proceedings of the IEEE Conference on Computer Vision and
  Pattern Recognition}, volume~1, pages 419--426, 2006.

\bibitem{tang2013content}
Xiaoou Tang, Wei Luo, and Xiaogang Wang.
\newblock Content-based photo quality assessment.
\newblock {\em IEEE Transactions on Multimedia}, 15(8):1930--1943, 2013.

\bibitem{bhattacharya2010framework}
Subhabrata Bhattacharya, Rahul Sukthankar, and Mubarak Shah.
\newblock A framework for photo-quality assessment and enhancement based on
  visual aesthetics.
\newblock In {\em Proceedings of the 18th ACM International Conference on
  Multimedia}, pages 271--280. ACM, 2010.

\bibitem{dhar2011high}
S~Dhar, V~Ordonez, and TL~Berg.
\newblock High level describable attributes for predicting aesthetics and
  interestingness.
\newblock In {\em Proceedings of the 2011 IEEE Conference on Computer Vision
  and Pattern Recognition}, pages 1657--1664, 2011.

\bibitem{su2011scenic}
Hsiao-Hang Su, Tse-Wei Chen, Chieh-Chi Kao, Winston~H Hsu, and Shao-Yi Chien.
\newblock Scenic photo quality assessment with bag of aesthetics-preserving
  features.
\newblock In {\em Proceedings of the 19th ACM International Conference on
  Multimedia}, pages 1213--1216. ACM, 2011.

\bibitem{li2010towards}
Congcong Li, Alexander~C Loui, and Tsuhan Chen.
\newblock Towards aesthetics: A photo quality assessment and photo selection
  system.
\newblock In {\em Proceedings of the 18th ACM International Conference on
  Multimedia}, pages 827--830. ACM, 2010.

\bibitem{he2016deep}
Kaiming He, Xiangyu Zhang, Shaoqing Ren, and Jian Sun.
\newblock Deep residual learning for image recognition.
\newblock In {\em Proceedings of the IEEE Conference on Computer Vision and
  Pattern Recognition}, pages 770--778, 2016.

\bibitem{szegedy2016rethinking}
Christian Szegedy, Vincent Vanhoucke, Sergey Ioffe, Jon Shlens, and Zbigniew
  Wojna.
\newblock Rethinking the inception architecture for computer vision.
\newblock In {\em Proceedings of the IEEE Conference on Computer Vision and
  Pattern Recognition}, pages 2818--2826, 2016.

\bibitem{murray2012ava}
Naila Murray, Luca Marchesotti, and Florent Perronnin.
\newblock Ava: A large-scale database for aesthetic visual analysis.
\newblock In {\em Proceedings of the IEEE Conference on Computer Vision and
  Pattern Recognition}, pages 2408--2415, 2012.

\bibitem{Lu2015DeepMA}
Xin Lu, Zhe Lin, Xiaohui Shen, Radomir Mech, and James~Z Wang.
\newblock Deep multi-patch aggregation network for image style, aesthetics, and
  quality estimation.
\newblock In {\em Proceedings of the IEEE International Conference on Computer
  Vision}, pages 990--998, 2015.

\bibitem{Ma2017ALampAL}
Shuang Ma, Jing Liu, and Chang Wen~Chen.
\newblock A-lamp: Adaptive layout-aware multi-patch deep convolutional neural
  network for photo aesthetic assessment.
\newblock In {\em Proceedings of the IEEE Conference on Computer Vision and
  Pattern Recognition}, pages 4535--4544, 2017.

\bibitem{long2015fully}
Jonathan Long, Evan Shelhamer, and Trevor Darrell.
\newblock Fully convolutional networks for semantic segmentation.
\newblock In {\em Proceedings of the IEEE Conference on Computer Vision and
  Pattern Recognition}, pages 3431--3440, 2015.

\bibitem{he2015spatial}
Kaiming He, Xiangyu Zhang, Shaoqing Ren, and Jian Sun.
\newblock Spatial pyramid pooling in deep convolutional networks for visual
  recognition.
\newblock {\em IEEE Transactions on Pattern Analysis and Machine Intelligence},
  37(9):1904--1916, 2015.

\bibitem{fang2018image}
Huidi Fang, Chaoran Cui, Xiang Deng, Xiushan Nie, Muwei Jian, and Yilong Yin.
\newblock Image aesthetic distribution prediction with fully convolutional
  network.
\newblock In {\em International Conference on Multimedia Modeling}, pages
  267--278. Springer, 2018.

\bibitem{apostolidis2019image}
Konstantinos Apostolidis and Vasileios Mezaris.
\newblock Image aesthetics assessment using fully convolutional neural
  networks.
\newblock In {\em International Conference on Multimedia Modeling}, pages
  361--373. Springer, 2019.

\bibitem{cui2018distribution}
Chaoran Cui, Huihui Liu, Tao Lian, Liqiang Nie, Lei Zhu, and Yilong Yin.
\newblock Distribution-oriented aesthetics assessment with semantic-aware
  hybrid network.
\newblock {\em IEEE Transactions on Multimedia}, 21(5):1209--1220, 2018.

\bibitem{girshick2015fast}
Ross Girshick.
\newblock Fast r-cnn.
\newblock In {\em Proceedings of the IEEE International Conference on Computer
  Vision}, pages 1440--1448, 2015.

\bibitem{zeng2020grid}
Hui Zeng, Lida Li, Zisheng Cao, and Lei Zhang.
\newblock Grid anchor based image cropping: A new benchmark and an efficient
  model.
\newblock {\em IEEE Transactions on Pattern Analysis and Machine Intelligence},
  2020.

\bibitem{tu2020image}
Yi~Tu, Li~Niu, Weijie Zhao, Dawei Cheng, and Liqing Zhang.
\newblock Image cropping with composition and saliency aware aesthetic score
  map.
\newblock In {\em Proceedings of the AAAI Conference on Artificial
  Intelligence}, pages 12104--12111, 2020.

\bibitem{ren2017personalized}
Jian Ren, Xiaohui Shen, Zhe Lin, Radomir Mech, and David~J Foran.
\newblock Personalized image aesthetics.
\newblock In {\em Proceedings of the IEEE International Conference on Computer
  Vision}, pages 638--647, 2017.

\bibitem{VotingGuidelines}
Voting guidelines.
\newblock \url{https://www.dpchallenge.com/challenge_rules.php?RULES_ID=13}.

\bibitem{vaswani2017attention}
Ashish Vaswani, Noam Shazeer, Niki Parmar, Jakob Uszkoreit, Llion Jones,
  Aidan~N Gomez, {\L}ukasz Kaiser, and Illia Polosukhin.
\newblock Attention is all you need.
\newblock In {\em Advances in Neural Information Processing Systems}, pages
  5998--6008, 2017.

\bibitem{jin2018predicting}
Xin Jin, Le~Wu, Xiaodong Li, Siyu Chen, Siwei Peng, Jingying Chi, Shiming Ge,
  Chenggen Song, and Geng Zhao.
\newblock Predicting aesthetic score distribution through cumulative
  jensen-shannon divergence.
\newblock In {\em Thirty-Second AAAI Conference on Artificial Intelligence},
  pages 77--84, 2018.

\bibitem{kao2016hierarchical}
Yueying Kao, Kaiqi Huang, and Steve Maybank.
\newblock Hierarchical aesthetic quality assessment using deep convolutional
  neural networks.
\newblock {\em Signal Processing: Image Communication}, 47:500--510, 2016.

\bibitem{hosu2019effective}
Vlad Hosu, Bastian Goldlucke, and Dietmar Saupe.
\newblock Effective aesthetics prediction with multi-level spatially pooled
  features.
\newblock In {\em Proceedings of the IEEE Conference on Computer Vision and
  Pattern Recognition}, pages 9375--9383, 2019.

\bibitem{lee2019image}
Jun-Tae Lee and Chang-Su Kim.
\newblock Image aesthetic assessment based on pairwise comparison a unified
  approach to score regression, binary classification, and personalization.
\newblock In {\em Proceedings of the IEEE International Conference on Computer
  Vision}, pages 1191--1200, 2019.

\bibitem{talebi2018nima}
Hossein Talebi and Peyman Milanfar.
\newblock Nima: Neural image assessment.
\newblock {\em IEEE Transactions on Image Processing}, 27(8):3998--4011, 2018.

\bibitem{zhang2019gated}
Xiaodan Zhang, Xinbo Gao, Wen Lu, and Lihuo He.
\newblock A gated peripheral-foveal convolutional neural network for unified
  image aesthetic prediction.
\newblock {\em IEEE Transactions on Multimedia}, 21(11):2815--2826, 2019.

\bibitem{freund1997decision}
Yoav Freund and Robert~E Schapire.
\newblock A decision-theoretic generalization of on-line learning and an
  application to boosting.
\newblock {\em Journal of Computer and System Sciences}, 55(1):119--139, 1997.

\bibitem{he2009robust}
Ran He, Bao-Gang Hu, and Xiao-Tong Yuan.
\newblock Robust discriminant analysis based on nonparametric maximum entropy.
\newblock In {\em Asian Conference on Machine Learning}, pages 120--134.
  Springer, 2009.

\bibitem{luo2008photo}
Yiwen Luo and Xiaoou Tang.
\newblock Photo and video quality evaluation: Focusing on the subject.
\newblock In {\em European Conference on Computer Vision}, pages 386--399.
  Springer, 2008.

\bibitem{oliva2001modeling}
Aude Oliva and Antonio Torralba.
\newblock Modeling the shape of the scene: A holistic representation of the
  spatial envelope.
\newblock {\em International Journal of Computer Vision}, 42(3):145--175, 2001.

\bibitem{lowe2004distinctive}
David~G Lowe.
\newblock Distinctive image features from scale-invariant keypoints.
\newblock {\em International Journal of Computer Vision}, 60(2):91--110, 2004.

\bibitem{marchesotti2011assessing}
Luca Marchesotti, Florent Perronnin, Diane Larlus, and Gabriela Csurka.
\newblock Assessing the aesthetic quality of photographs using generic image
  descriptors.
\newblock In {\em Proceedings of the IEEE International Conference on Computer
  Vision}, pages 1784--1791, 2011.

\bibitem{song2019geometry}
Linsen Song, Jie Cao, Lingxiao Song, Yibo Hu, and Ran He.
\newblock Geometry-aware face completion and editing.
\newblock In {\em Proceedings of the AAAI Conference on Artificial
  Intelligence}, pages 2506--2513, 2019.

\bibitem{fu2021high}
Chaoyou Fu, Yibo Hu, Xiang Wu, Guoli Wang, Qian Zhang, and Ran He.
\newblock High-fidelity face manipulation with extreme poses and expressions.
\newblock {\em IEEE Transactions on Information Forensics and Security},
  16:2218--2231, 2021.

\bibitem{ma2021contrastive}
Xin Ma, Xiaoqiang Zhou, Huaibo Huang, Gengyun Jia, Zhenhua Chai, and Xiaolin
  Wei.
\newblock Contrastive attention network with dense field estimation for face
  completion.
\newblock {\em Pattern Recognition}, page 108465, 2021.

\bibitem{lu2014rapid}
Xin Lu, Zhe Lin, Hailin Jin, Jianchao Yang, and James~Z Wang.
\newblock Rapid: Rating pictorial aesthetics using deep learning.
\newblock In {\em Proceedings of the 22nd ACM International Conference on
  Multimedia}, pages 457--466. ACM, 2014.

\bibitem{lu2015rating}
Xin Lu, Zhe Lin, Hailin Jin, Jianchao Yang, and James~Z Wang.
\newblock Rating image aesthetics using deep learning.
\newblock {\em IEEE Transactions on Multimedia}, 17(11):2021--2034, 2015.

\bibitem{kao2015visual}
Yueying Kao, Chong Wang, and Kaiqi Huang.
\newblock Visual aesthetic quality assessment with a regression model.
\newblock In {\em Proceedings of the IEEE International Conference on Image
  Processing}, pages 1583--1587, 2015.

\bibitem{dong2015photo}
Zhe Dong, Xu~Shen, Houqiang Li, and Xinmei Tian.
\newblock Photo quality assessment with dcnn that understands image well.
\newblock In {\em International Conference on Multimedia Modeling}, pages
  524--535. Springer, 2015.

\bibitem{tian2015query}
Xinmei Tian, Zhe Dong, Kuiyuan Yang, and Tao Mei.
\newblock Query-dependent aesthetic model with deep learning for photo quality
  assessment.
\newblock {\em IEEE Transactions on Multimedia}, 17(11):2035--2048, 2015.

\bibitem{kao2017deep}
Yueying Kao, Ran He, and Kaiqi Huang.
\newblock Deep aesthetic quality assessment with semantic information.
\newblock {\em IEEE Transactions on Image Processing}, 26(3):1482--1495, 2017.

\bibitem{wang2019aspect}
Lijie Wang, Xueting Wang, Toshihiko Yamasaki, and Kiyoharu Aizawa.
\newblock Aspect-ratio-preserving multi-patch image aesthetics score
  prediction.
\newblock In {\em Proceedings of the IEEE Conference on Computer Vision and
  Pattern Recognition Workshops}, pages 0--0, 2019.

\bibitem{sheng2018attention}
Kekai Sheng, Weiming Dong, Chongyang Ma, Xing Mei, Feiyue Huang, and Bao-Gang
  Hu.
\newblock Attention-based multi-patch aggregation for image aesthetic
  assessment.
\newblock In {\em Proceedings of the 26th ACM International Conference on
  Multimedia}, pages 879--886, 2018.

\bibitem{Mai2016CompositionPreservingDP}
Long Mai, Hailin Jin, and Feng Liu.
\newblock Composition-preserving deep photo aesthetics assessment.
\newblock In {\em Proceedings of the IEEE Conference on Computer Vision and
  Pattern Recognition}, pages 497--506, 2016.

\bibitem{chen2020adaptive}
Qiuyu Chen, Wei Zhang, Ning Zhou, Peng Lei, Yi~Xu, Yu~Zheng, and Jianping Fan.
\newblock Adaptive fractional dilated convolution network for image aesthetics
  assessment.
\newblock In {\em Proceedings of the IEEE/CVF Conference on Computer Vision and
  Pattern Recognition}, pages 14114--14123, 2020.

\bibitem{li2020personality}
Leida Li, Hancheng Zhu, Sicheng Zhao, Guiguang Ding, and Weisi Lin.
\newblock Personality-assisted multi-task learning for generic and personalized
  image aesthetics assessment.
\newblock {\em IEEE Transactions on Image Processing}, 29:3898--3910, 2020.

\bibitem{wu2011learning}
Ou~Wu, Weiming Hu, and Jun Gao.
\newblock Learning to predict the perceived visual quality of photos.
\newblock In {\em Proceedings of the IEEE International Conference on Computer
  Vision}, pages 225--232, 2011.

\bibitem{jin2016image}
Bin Jin, Maria V~Ortiz Segovia, and Sabine S{\"u}sstrunk.
\newblock Image aesthetic predictors based on weighted cnns.
\newblock In {\em 2016 IEEE International Conference on Image Processing},
  pages 2291--2295. Ieee, 2016.

\bibitem{he2017mask}
Kaiming He, Georgia Gkioxari, Piotr Doll{\'a}r, and Ross Girshick.
\newblock Mask r-cnn.
\newblock In {\em Proceedings of the IEEE International Conference on Computer
  Vision}, pages 2961--2969, 2017.

\bibitem{dai2016r}
Jifeng Dai, Yi~Li, Kaiming He, and Jian Sun.
\newblock R-fcn: Object detection via region-based fully convolutional
  networks.
\newblock In {\em Advances in Neural Information Processing Dystems}, pages
  379--387, 2016.

\bibitem{lin2017feature}
Tsung-Yi Lin, Piotr Doll{\'a}r, Ross Girshick, Kaiming He, Bharath Hariharan,
  and Serge Belongie.
\newblock Feature pyramid networks for object detection.
\newblock In {\em Proceedings of the IEEE Conference on Computer Vision and
  Pattern Recognition}, pages 2117--2125, 2017.

\bibitem{cai2018cascade}
Zhaowei Cai and Nuno Vasconcelos.
\newblock Cascade r-cnn: Delving into high quality object detection.
\newblock In {\em Proceedings of the IEEE Conference on Computer Vision and
  Pattern Recognition}, pages 6154--6162, 2018.

\bibitem{challenge_history}
Challenge history.
\newblock \url{https://www.dpchallenge.com/challenge_history.php}.

\bibitem{datta2008algorithmic}
Ritendra Datta, Jia Li, and James~Z Wang.
\newblock Algorithmic inferencing of aesthetics and emotion in natural images:
  An exposition.
\newblock In {\em 2008 15th IEEE International Conference on Image Processing},
  pages 105--108. IEEE, 2008.

\bibitem{kong2016photo}
Shu Kong, Xiaohui Shen, Zhe Lin, Radomir Mech, and Charless Fowlkes.
\newblock Photo aesthetics ranking network with attributes and content
  adaptation.
\newblock In {\em European Conference on Computer Vision}, pages 662--679.
  Springer, 2016.

\bibitem{meng2018mlans}
Xuantong Meng, Fei Gao, Shengjie Shi, Suguo Zhu, and Jingjie Zhu.
\newblock Mlans: Image aesthetic assessment via multi-layer aggregation
  networks.
\newblock In {\em 2018 Eighth International Conference on Image Processing
  Theory, Tools and Applications (IPTA)}, pages 1--6. IEEE, 2018.

\bibitem{devlin2018bert}
Jacob Devlin, Ming-Wei Chang, Kenton Lee, and Kristina Toutanova.
\newblock Bert: Pre-training of deep bidirectional transformers for language
  understanding.
\newblock {\em arXiv preprint arXiv:1810.04805}, 2018.

\bibitem{wolf-etal-2020-transformers}
Thomas Wolf, Lysandre Debut, Victor Sanh, Julien Chaumond, Clement Delangue,
  Anthony Moi, Pierric Cistac, Tim Rault, Rémi Louf, Morgan Funtowicz, Joe
  Davison, Sam Shleifer, Patrick von Platen, Clara Ma, Yacine Jernite, Julien
  Plu, Canwen Xu, Teven~Le Scao, Sylvain Gugger, Mariama Drame, Quentin Lhoest,
  and Alexander~M. Rush.
\newblock Transformers: State-of-the-art natural language processing.
\newblock In {\em Proceedings of the 2020 Conference on Empirical Methods in
  Natural Language Processing: System Demonstrations}, pages 38--45, Online,
  October 2020. Association for Computational Linguistics.

\end{thebibliography}

\begin{IEEEbiography}[{\includegraphics[width=1in,height=1.25in,clip,keepaspectratio]{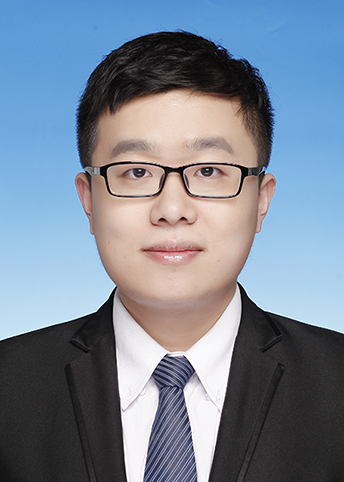}}]{Gengyun Jia}
received the B.E. degree in communication engineering from Shandong University (SDU), Jinan, China, in 2015, and the M.S. degree in information and communication engineering from Beijing University of Posts and Telecommunications (BUPT), Beijing, China, in 2018. He is currently pursuing the Ph.D degree in computer application technology with the University of Chinese Academy of Sciences (UCAS), Beijing, China, and with the National Laboratory of Pattern Recognition, Center for Research on Intelligent Perception and Computing, Institute of Automation, Chinese Academy of Sciences, Beijing, China. His research interests include image understanding, media forensics and machine learning.
\end{IEEEbiography}

\begin{IEEEbiography}[{\includegraphics[width=1in,height=1.25in,clip,keepaspectratio]{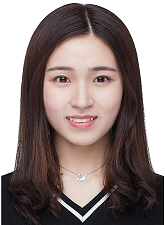}}]{Peipei Li}
received the B.S. degree from the Information and Control Engineering of China University of Petroleum in 2016, and the M.S. and Ph.D. degree in pattern recognition and intelligent systems from the Chinese Academy of Sciences, Beijing, China, in 2021. Since 2021, she has been a Faculty Member with the School of Artificial Intelligence, Beijing University of Posts and Telecommunications, Beijing 100876, China, where she is currently an Associate Professor. Her current research interests include deep learning, computer vision, biometrics, and machine learning.
\end{IEEEbiography}

\begin{IEEEbiography}[{\includegraphics[width=1in,height=1.25in,clip,keepaspectratio]{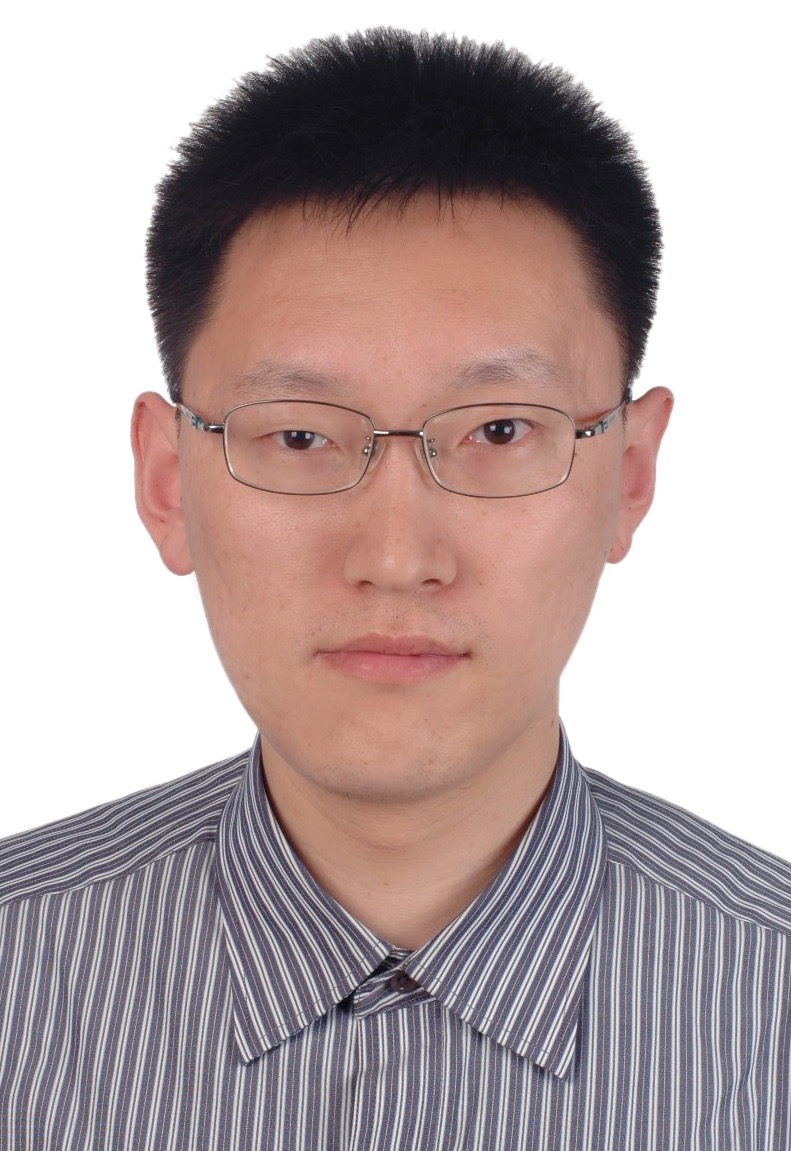}}]{Ran He}
received the BE and MS degrees in computer science from the Dalian University of Technology, Dalian, China, 2001 and 2004, respectively, and the PhD degree in pattern recognition and intelligent systems from the Institute of Automation, Chinese Academy of Sciences (CASIA), Beijing, China, in 2009. Since September 2010, he has joined NLPR where he is currently a full professor. He serves as the editor board member of IEEE TIP and PR, and serves on the program committee of several conferences. His research interests focus on information theoretic learning, pattern recognition, and computer vision. He is a senior member of the IEEE and the Fellow of the IAPR.
\end{IEEEbiography}




%








\end{document}